\documentclass[wcp]{jmlr}

\usepackage{longtable}
\usepackage{rotating}

\usepackage{booktabs}
\usepackage{graphicx}
\usepackage[utf8]{inputenc} 
\usepackage[T1]{fontenc}    
\usepackage{hyperref}       
\usepackage{url}            
\usepackage{booktabs}       
\usepackage{amsfonts}       
\usepackage{nicefrac}       
\usepackage{microtype}      
\usepackage{xcolor}         
\usepackage{amsmath,amssymb,amsfonts}
\usepackage{threeparttable}
\usepackage{wrapfig}
\usepackage{graphicx}
\usepackage{textcomp}
\usepackage{xcolor}
\usepackage{natbib}
\setcitestyle{square, comma, numbers,sort&compress, super}
\usepackage{url}
\usepackage{multirow}
\usepackage{booktabs}
\usepackage{stfloats}
\usepackage{graphicx}
\usepackage{array}
\usepackage{diagbox}
\usepackage{caption}  
\usepackage{hyperref}
\usepackage{mathrsfs}   
\usepackage[T1]{fontenc}
\usepackage{threeparttable}         
\usepackage{algorithm}
\usepackage{algorithmicx}
\usepackage{algpseudocode}
\usepackage{lipsum}
\usepackage{graphicx}
\usepackage{color}
\usepackage{pifont}
\usepackage{algpseudocode}
\usepackage{multirow}
\usepackage{soul}
\usepackage{threeparttable}
\usepackage{physics}
\usepackage{amsmath}
\usepackage{tikz}
\usepackage{mathdots}
\usepackage{yhmath}
\usepackage{cancel}
\usepackage{color}
\usepackage{siunitx}
\usepackage{array}
\usepackage{multirow}
\usepackage{amssymb}
\usepackage{gensymb}
\usepackage{extarrows}
\usepackage{booktabs}
\usetikzlibrary{fadings}
\usetikzlibrary{patterns}
\usetikzlibrary{shadows.blur}
\usetikzlibrary{shapes}

\graphicspath{{properties/}}   
\floatname{algorithm}{Algorithm}

\usepackage{lineno}

\makeatletter
\let\Ginclude@graphics\@org@Ginclude@graphics 
\makeatother

\jmlrvolume{222}
\jmlryear{2023}
\jmlrworkshop{ACML 2023}

\title[Spiking Neural Network with Minimum Latency]{Training a General Spiking Neural Network with Improved Efficiency and Minimum Latency}

\author{\Name{Yunpeng Yao} \Email{202132748@mail.sdu.edu.cn}\\
  \addr Information Science and Engineering, Shandong University, Shandong, China
  \AND
  \Name{Man Wu\textsuperscript{$\ast$}} \Email{wu.man.wi5@am.ics.keio.ac.jp}\\
 \addr Department of Information and Computer Science, Keio University, Kanagawa, Japan 
   \AND
  \Name{Zheng	Chen} \Email{chenz@sanken.osaka-u.ac.jp}\\
 \addr SANKEN, Osaka University, Osaka, Japan
   \AND
  \Name{Renyuan Zhang} \Email{rzhang@is.naist.jp}\\
 \addr Division of Information Science, Nara Institute of Science and Technology, Nara, Japan \\
\addr $\ast$ Corresponding Author
 }

\editors{Berrin Yan{\i}ko\u{g}lu and Wray Buntine}

\begin{document}

\maketitle

\begin{abstract}
Spiking Neural Networks (SNNs) that operate in an event-driven manner and employ binary spike representation have recently emerged as promising candidates for energy-efficient computing.
However, a cost bottleneck arises in obtaining high-performance SNNs: training a SNN model requires a large number of time steps in addition to the usual learning iterations, hence this limits their energy efficiency.
This paper proposes a general training framework that enhances feature learning and activation efficiency within a limited time step, providing a new solution for more energy-efficient SNNs. 
Our framework allows SNN neurons to learn robust spike feature from different receptive fields and update neuron states by utilizing both current stimuli and recurrence information transmitted from other neurons.
This setting continuously complements information within a single time step.
Additionally, we propose a projection function to merge these two stimuli to smoothly optimize neuron weights (spike firing threshold and activation).
We evaluate the proposal for both convolution and recurrent models.
Our experimental results indicate state-of-the-art visual classification tasks, including CIFAR10, CIFAR100, and TinyImageNet. 
Our framework achieves 72.41$\%$ and 72.31$\%$ top-1 accuracy with only $1$ time step on CIFAR100 for CNNs and RNNs, respectively.
Our method reduces $10\times$ and $3\times$ joule energy than a standard ANN and SNN, respectively, on CIFAR10, without additional time steps.
\end{abstract}

\section{Introduction}

Spiking Neural Networks (SNNs) have garnered attention due to the capacity of low-power computing, which is inspired by the event-driven nature of human brain.
The neurons in SNNs operate by transmitting information through event-oriented \emph{binary spike} as opposed to analog value, mitigating the computational cost associated with multiplication operations that are ubiquitous in ANNs \cite{AAAI-2023-Peter, hpcc22}.

Despite their notable energy efficiency, high-performance SNNs are notoriously difficult to develop, particularly for continuous and high-dimensional signals \cite{ECCV-2022-roy}.
SNN neurons utilize an accumulating-and-firing mechanism, known as the membrane potential, which is a dynamic variable that evolves over time, unlike the fixed weights.
This mechanism allows neurons to receive and transmit information through activated spike sequences.
However, guiding representative activation for continuous input is tough.
The sparsity that arises from discrete spike representation inherently results in a consequent loss of information \cite{AAAI-2023-Peter}, making SNNs more difficult for learning complex patterns in real-world signals.
Meanwhile, the discrete nature of spikes also poses an obstacle in model convergence, leading to a gradual decay of effective spike activation in deeper layers \cite{TNN-2021-roy}.

SNNs trained using the ANN-to-SNN conversion, which involves transferring the trained weights of an ANN onto an SNN, is a dominant method in the field \cite{rueckauer2017conversion}.
Coupled with appropriate learning algorithms, e.g., surrogate gradient learning (SGL) \cite{neftci2019surrogate}, conversion SNNs can perform comparable results to state-of-the-art (SOTA) ANNs methods \cite{cao2015spiking}.
Since additional training phases and meticulous attention of modeling are required \cite{simonyan2014very}, and computational complexity is inevitably increased.
Additionally, conversion SNNs typically cooperate with pre-encoding techniques (e.g., rate coding, rank-order coding, etc) to transform analog input into well-initialized spike sequences \cite{kiselev2016rate}.
These coding methods are iteratively applied over a large number of \emph{time steps} \footnote{time steps refer to the times of forward propagation in each spiking neuron.} for training, enabling continuous complementation of information source and optimizing membrane potential variable. 
Computationally, the time steps (usually >100) lead to a queuing of serial processing that increases end-to-end latency (proportional to the number of time steps) and costly memory \cite{AAAI-2023-Peter}.

To tackle this cost bottleneck, some works aim to develop a direct training framework that integrates the advanced computational mechanism to improve the optimization of SNN neurons.
For instance, strengthening the local feature receptive fields by convolutional neural networks (CNNs), some works reduce 50\% time steps required to achieve satisfactory results \cite{ ICLR-2022-Rathi}.
Recent studies employ hierarchical computation in CNN-based architectures, such as VGG and ResNet, within SNN frameworks \cite{kim2018deep}, which can train very deeper SNNs.
More recent studies \cite{ponghiran2022spiking} utilize the sequential-order capabilities of recurrent neural networks (RNNs) with SNNs to improve the recurrence dynamics for sequential learning. 
These investigations show efficacy with fewer time steps (< 20), which can result in improved computational efficiency \cite{ponghiran2022spiking}.
While the use of advanced computational mechanisms has yielded various SOTA results to SNNs research and reduced certain computational costs, it is still challenging to maintain accuracy in the direct training method as the number of time steps decreases. 
Typically, a reduction in time steps leads to a drop in accuracy, it leads to a \emph{trade-off between accuracy and the time steps used}. 
This trade-off issue imposes a limit on fully exploiting the low-power and energy-efficient computing capabilities of SNNs.
This paper provides a new perspective on this issue by proposing a novel and general learning framework for the direct training method.

Our focus is on maximizing the effective utilization of information in SNN neurons within limited time steps.
To achieve this, we propose a learning framework that involves (i) learning feature from different input receptive fields and (ii) optimizing the spike stimuli through a projection function.
Specifically, SNN neurons receive multiple local inputs, which are extracted using a sliding window approach and grouped accordingly in each layer. 
This enables SNN neurons to learn information at different levels of granularity to model dependencies at different scales.
We record the group-wise membrane potentials and recurrently utilize them to optimize SNN neurons within a single time step. 
We further propose a projection function to smooth neuron activation, optimizing the membrane potential by both previous group membrane potential and current stimuli. This facilitates layer communication and ensures the effective transmission of information. The main contributions are summarized as follows:
\begin{itemize}
    \item This study tackles the trade-off between \emph{time-step} and \emph{performance}, and proposes a novel training framework for SNNs that involves the refinement of input information and the optimization of spiking neurons to enhance effective spike activation.
    \item Our framework incorporates a projection function to generalize its applicability to both Convolutional - and Recurrent- based architectures, based on an analysis of the activated operation of leaky integrate-and-fire (LIF) neurons. 
    \item Our method shows high effectiveness and efficiency in experimental results. 
    Our framework achieves top-1 accuracy of $72.31\%$ on CIFAR100, with only \texttt{1} time step. 
    Moreover, to the best of our knowledge, this is the first RNN-SNN model to achieve competitive results in visual tasks. On CIFAR10, our SNN significantly reduces computational cost with $(10\sim102)\times$ and $(1.4\sim3)\times$ joules compared to previous ANN and SNN approaches respectively.

\end{itemize}

\section{Preliminary: Leaky Integrate and Fired Model}
SNNs are computationally capable of universal approximation and can mimic any feed-forward neural network by adjusting synaptic weights. Thus, they excel at processing spatio-temporal and fluctuation information as spike flows. 

There have been many proposals for different types of SNN neurons, in this work,
we follow the works of \cite{lotfi2020long, ChenIJCAI, AAAI-2023-Peter} and use the well-known iterative leaky integrate-and-fire (LIF) SNN model in this work.
The conventional LIF describes spiking nature of neurons from feature stimuli, membrane potential accumulation, and spike firing, to membrane potential reset as membrane current $I(t)$ and membrane potential $V(t)$.
Neuronal dynamics can be described by

\begin{equation}
{\displaystyle \frac{\partial I(t)}{\partial t} =}-\tau _{syn}I(t)+\sum _{j=1}^{J}(w_{j} ,X_{j} (t-1))\ 
\label{eq3.1}
\end{equation}

\begin{equation}
\frac{\partial V\ (t)}{\partial t} =-\tau _{mem} V\ (t)+\ I(t)\ -\delta o_{ l}^{t-1}
\label{eq3.2}
\end{equation}
where $t$ and $l$ denotes the $t$-th time-step and the $i$-th layer in the architecture, respectively. 
$V{(t)}$ is the corresponding membrane potential and $\tau$ is the decay rate of the current and potential. 
Hence, $o^{i}_{t-1}$ represents the $i$-th firing spike with weight $w_{j,n}$ from the previous layer and time step ($t-1$). 
Given the above information, $I(t)$ is the pre-synaptic input, and its information carrier is activated by the previous time index.
When the membrane potential $V(t)$ reaches the firing threshold $\delta$, the neuron will output a firing spike, i.e., \emph{1}.
\begin{equation}
o_{t+1}^i = 
\begin{cases}
    1 & \text{ if } V(t)>\delta \\
0  & \text{ otherwise }
\end{cases}
\label{eq3.3}
\end{equation}

SNNs place less emphasis on the number of spikes indicated by $1$, and instead, consider the number of $0$ between two $1$ spikes as a reflection of the accumulation of event status, which is incorporated into information representations. The spike generated by $o_{t+1}^i$ propagates forward and activates the neurons of the next layer.

\section{Methodology}
The goal of this work is to enable SNNs to learn visual features without requiring a large number of time steps.
To achieve this, we use an input stream $X \in \mathbb{R}_{x}^{\left( G_{x}^{1},...,G_{x}^{P}\right) \times T}$ that consists of a binary signal ($0/1$), which is divided into $P$ groups up to time step $T$, with each group having a feature space of $G_{x}$.
Equally, we also use a membrane potential stream $M \in \mathbb{R}_{m}^{\left( G{m}^{1},...,G_{m}^{P}\right) \times T}$ that corresponds to $X_{l}$.
Based on these, the proposed SNN model $F\ (X|\beta;\Theta;T)$ aims to fit the target distribution $Y$, where $\Theta$ is a set of model parameters (such as weight $w_{j,n}$ from the previous layer) and $\beta$ is an auxiliary parameter that describes membrane potential of SNN. 
In this sense, we propose a novel framework to optimize the learning ability of $F$ with a minimum iteration time of $T$.
\begin{figure}[t!]
\centerline{\includegraphics[width=0.75\linewidth]{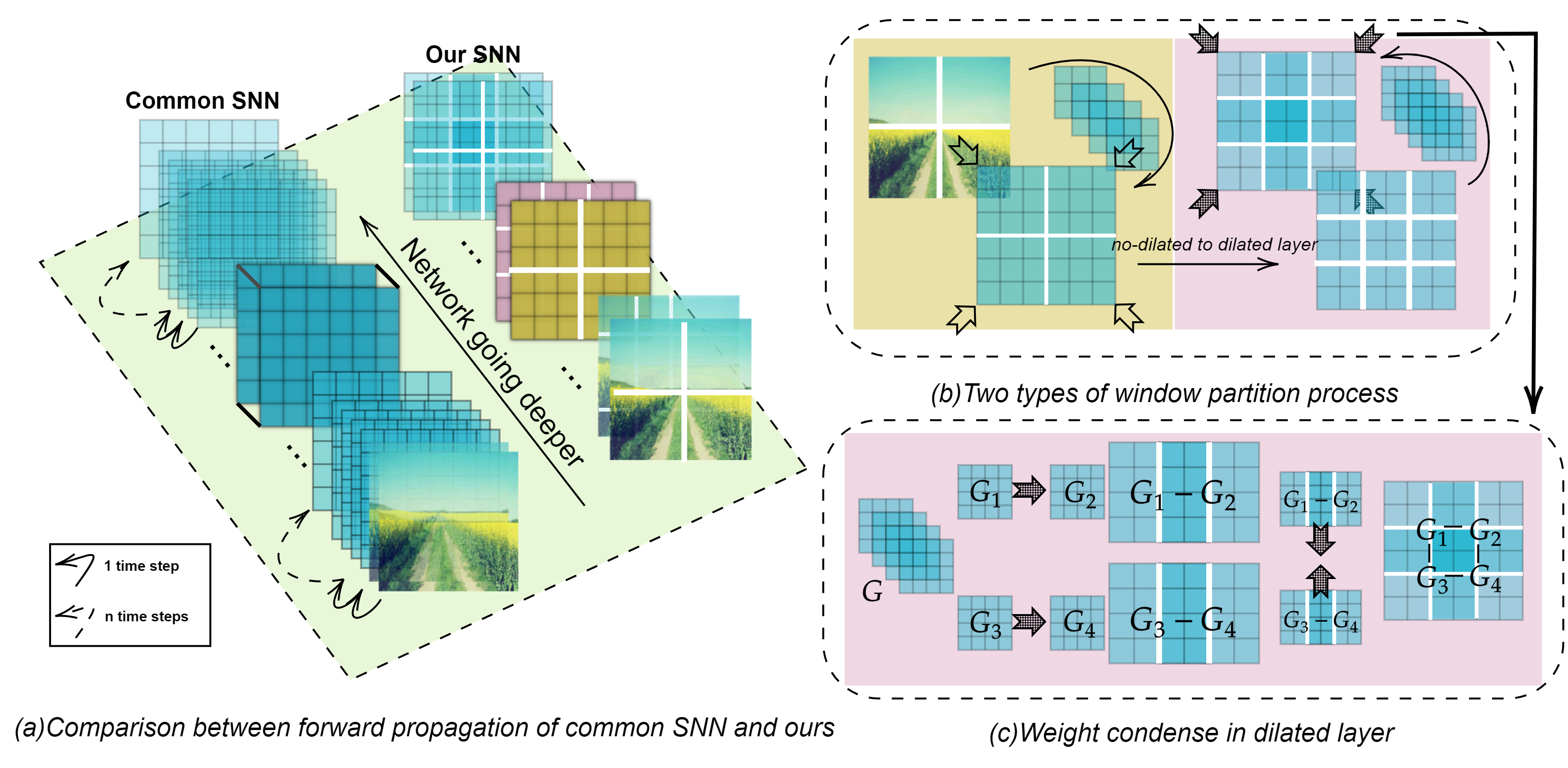}}
\caption{Windows partition for spiking neuron layers.}
\label{window}
\end{figure}

\subsection{Dilated Window for Spiking Neuron.}
\label{section3.1}
Inspired by the works of Dosovitskiy et al. \cite{dosovitskiy2020image} and Liu et al. \cite{liu2021swin}, which regionalize global information, our approach employs complementary paired-windows. 
However, we modify one of them to accommodate the operation of spiking neurons between consecutive layers as shown in Fig. \ref{window}(b).
Specifically, non-dilated windows are designed to accumulate membrane potential locally, and dilated windows are proposed to enhance the interaction between neighboring windows. 
\par

\textbf{Membrane potential in non-dilated windows.} 
The non-dilated windows are arranged to evenly partition the membrane potential in a non-overlapping manner.
Under this condition, the tensor stream of input feature $X_{l}$ and membrane potential $M_{l}$ are first partitioned into $p$ groups, denoted as $G_{x}$ and $G_{m}$ respectively.
Then, each group spiking neuron $G_{x}$ accumulates membrane potential, activates a spike, and resets in order.
The accumulated membrane potential in $G_{x}^{p}$ will be delivered to the next $G_{x}^{p+1}$ in sequence to introduce cross-neuron connections up to iteration $T$.
Finally, every $G_{x}^{p}$ is activated by the membrane potential $G_{m}^{p-1}$ of last group and the weight $w$ corresponding to $G_{x}^{p}$, and $G_{x}^{1,...,p}$ is recomposed into origin shape as $X_{l}$.
Layers with partition windows can be represented as follow:
\begin{equation}
F_{l}( X|\beta ;\Theta ;t) =F_{l}\left( G_{x}^{1} |G_{m}^{1} ;\Theta ;t\right)\rightarrow F_{l}\left( G_{x}^{2} |G_{m}^{2} ;\Theta ;t\right)\xrightarrow{\dotsc } F_{l}\left( G_{x}^{p} |G_{m}^{p} ;\Theta ;t\right)
\label{eq4.2.1}
\end{equation}
 \par

\textbf{Membrane potential in dilated windows.}
The window-based LIF module mentioned earlier has a limitation in its modeling power, as it lacks connectivity across regions.
To address this issue, we propose a dilated window that alternates between two partitioning configurations in consecutive SNN blocks, thereby enhancing connectivity across windows.
The dilated window size dilates $1/2$ compared to the normal window, and the updating scheme of membrane potential is consistent with the normal window.
Each dilated window is therefore able to interact with all neighboring non-overlapping windows in the preceding layer as it expands the scope of partitioning.
Notice that the size of $G_{x}$ and $G_{m}$ partitioned by the dilated window is larger than origin $X_{l-1}$ and $M_{l-1}$ from non-dilated window after recomposing.
Consequently, the overlapping information will be overutilized since more additional feature space, and calculations will also be introduced.

To tackle this issue, we propose a more efficient computation approach that condenses the overlapping part with a certain weight, as shown in Fig. \ref{window}. Weighted-condensed compresses the information streams $X_l$ and $M_l$ into the original feature size with certain weights, namely $4$ and $2$.
The weight assigned to a region increases as the number of overlaps increases.
More specifically, due to the division operation of weighted-condense, “median spikes (Ms)” (such as 0.25, 0.5, and 0.75) will be produced in overlaps region of recomposed $X_{l+1}$. 
For instance, if a spike appears in the overlapping area with weight $4$, it becomes a Ms as $0.25$.
To maintain the low power advantage of event-driven, we set a region threshold ($Th_R$) to integrate these Ms into spikes. 
This threshold is set to 0.1.
Pseudo code for reorganizing a dilated window is presented in Appendix.

Due to the locality of dilated windows, a fully connected layer is established after each dilated layer to complete the learning of global features.
We therefore reduced the number of layers in the network to maintain the same amount of parameters with other researches.

\subsection{Fusion for Multi-Direction Membrane Potential}

As mentioned above, spiking neurons update their own activation status $X_{l}$ partially from the last window region. Although the membrane potential has accumulated more information by utilizing the rest feature region in the same layer, the information in the original area is still underutilized. To further enhance the descriptive ability of $\beta$, the fusion of multi-directional membrane potentials is critical. The most common fusion method is to add two sets of membrane potentials. \par
\begin{wrapfigure}{r}{0.30\textwidth}
\centerline{\includegraphics[width=1\linewidth,height=1\linewidth]{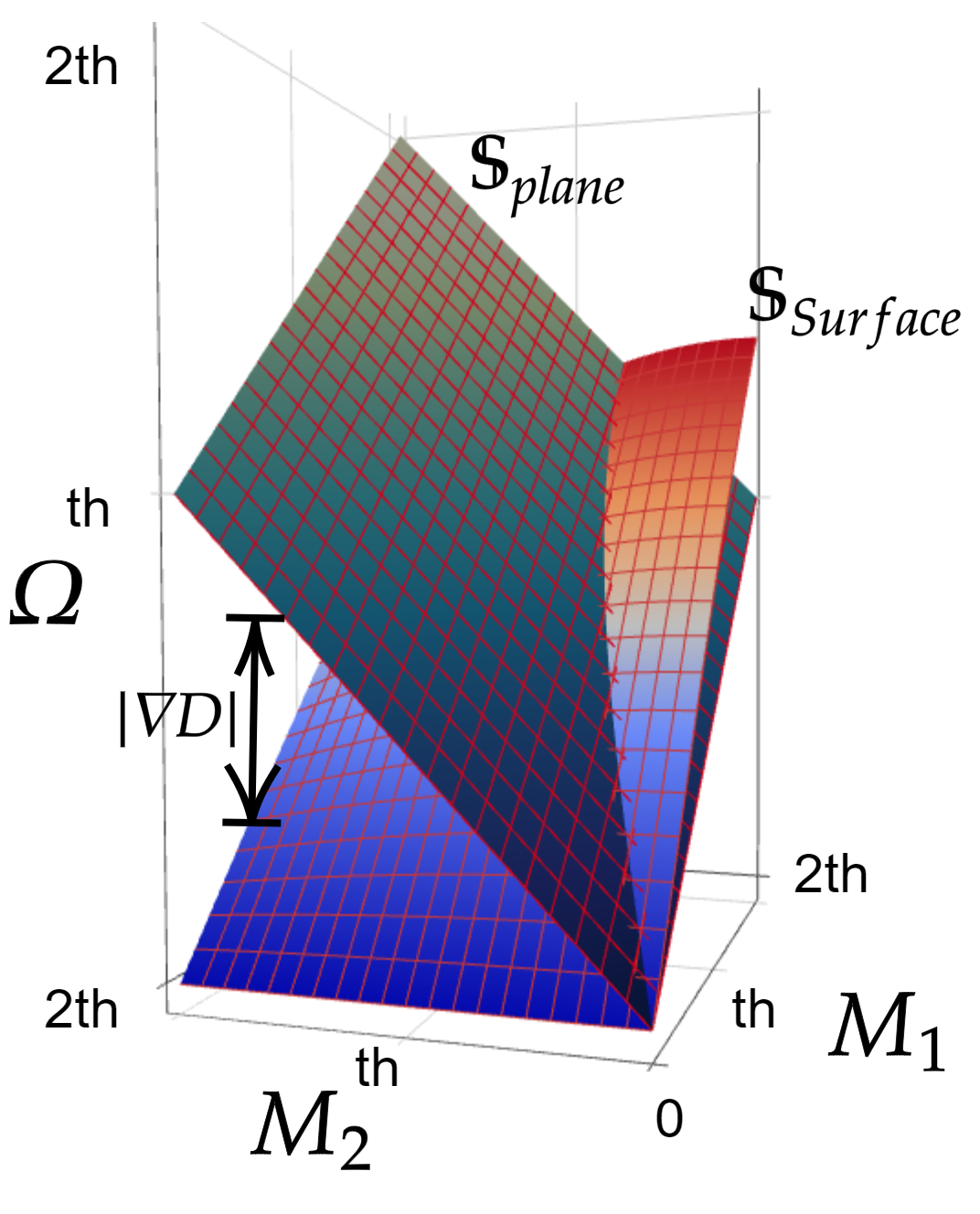}}
\caption{Example for $\theta=0.5$.}
\label{composefunc}
\end{wrapfigure}

\label{section3.2}
\begin{figure}[t]
    \centering
    \centerline{\includegraphics[width=0.75\linewidth]{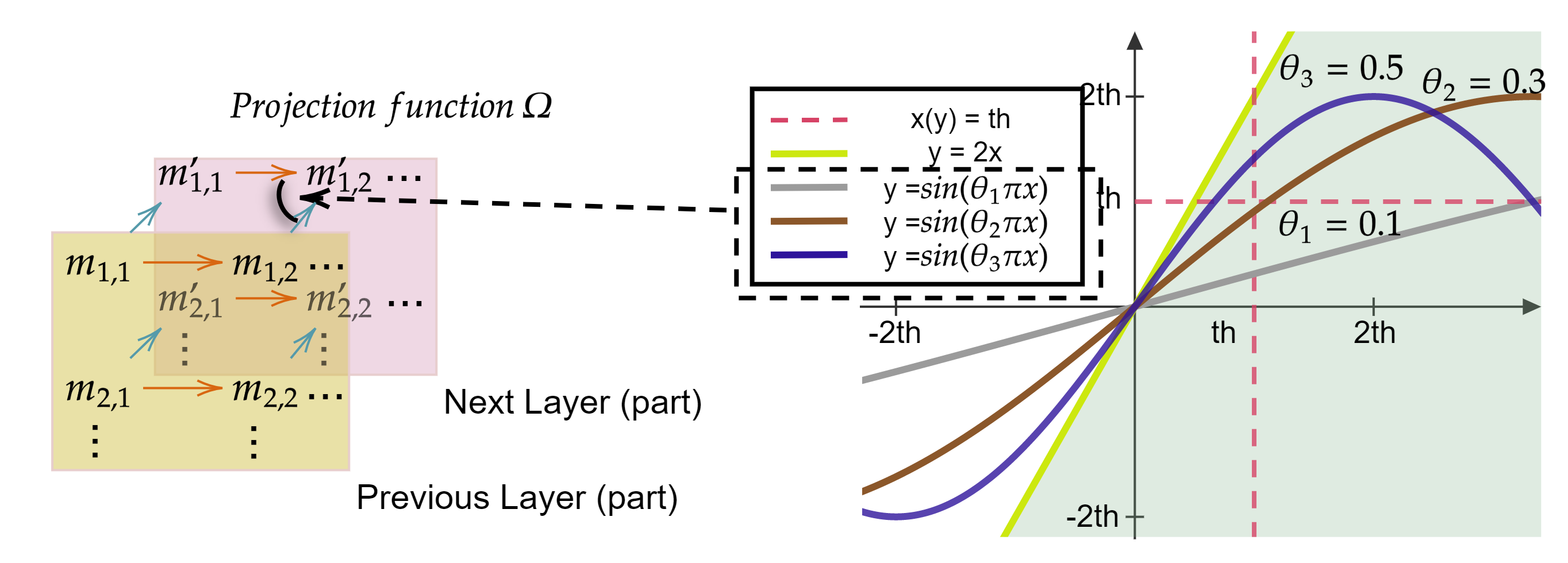}}
    \caption{Fusing membrane from two directions and projection function $\Omega$.}
    \label{fusion_method}
\end{figure}
As shown in Fig. \ref{fusion_method}, two issues emerge in this simplest fusion method (i.e., membrane potential streams $M$ in both directions are projected onto $y=2x$):
(i) Since $G_{m}$ always accumulates, including negative values that are not reset within one time step, it can lead to negative values accumulating towards infinity.
(ii) The simple linear sum function causes the fused membrane potential to increase monotonically and even exceed multiple LIF thresholds. This means that neurons in further regions will be over activated in advance without receiving any new information.
To fuse the membrane potential smoothly, we designed a projection function, denoted by $\Omega (\cdot)$, which projects the mixed membrane potential into an optimized space, as shown in Eq. \ref{eq4.1}: 
\begin{equation}
\label{eq4.1}
\Omega(G_{\beta}) =\Omega \left( G_{m},G_{m^{\prime }}\right) =2sin\left(\frac{\theta\pi }{2} G_{m}\right) \ *\ cos\left(\frac{\theta\pi }{2} G_{m^{\prime }}\right)
\end{equation}
where the variables $G_{m}$ and $G_{m^{\prime}}$ represent multi-directional membrane potentials from $G_{p-1}$ and $G_{l-1}$, respectively. Different values of $\theta$ correspond to different projection spaces. As shown in Fig. \ref{fusion_method}, Eq. \ref{eq4.1} deforms the function $y = \sin(\theta \pi x)$, going beyond simple linear summation of membrane potential from two directions. First, it effectively limits the trend toward negative infinity, allowing neurons in the negative area to be reactivated. Second, the membrane potential no longer exceeds multiple LIF thresholds, greatly reducing the influence of spiking neuron $G_{m}^{p,l}$ on the "remote future" $G_{x}^{P,L}$. 
Additionally, non-linearity is introduced, which smooths neuron activation and improves the representation of complex distributions by the neurons.
It is worth noting that the function $\Omega (\cdot)$ used for the projection is not necessarily the optimal expression for achieving multi-directional membrane potential fusion. Several other possible functions and different values of $\theta$ are experimented with in section 4.3.

where $G_{m}$ and $G_{m^{\prime }}$ present multi-direction membrane potential from $G_{p-1}$ and $G_{l-1}$ respectively.
And different $\theta$ represent different projection spaces. 
As Fig. \ref{composefunc} shown, Eq. \ref{eq4.1} is a deformation of $y = sin(\theta \pi x)$, which surpasses the simple linear summation of membrane potential from two directions.
First, it is obvious that the trend of negative infinity has been well restrained, that neurons in the negative area will have an opportunity to be reactivated. 
Second, membrane potential will not exceed multiple LIF thresholds anymore, limiting the influence of spiking neuron $G_{m}^{p,l}$ on the "remote future" $G_{x}^{P,L}$ to a great extent.
In addition, non-linearity is introduced that smooths neuron activation and improves neuronal representation of complex distributions.
As a note, it is not said that project function $\Omega ( \cdot )$ is the optimal expression for achieving multi-direction membrane potential fusion.
Several possible functions and different $\theta$ are experimented as ablation in section 5.3.
As shown in Fig.\ref{composefunc}, two issues will emerge in this simplest fusion way (i.e, membrane potential streams $M$ in both directions are projected on $y=x$ or $z=x+y$):
(i) Since $G_{m}$ will always accumulate including negative values and not reset (in 1-time step), including negative values in the direction of infinity.
(ii) The simple linear sum function makes the fused membrane potential monotonically increase and even exceed multiple LIF thresholds.
This means that neurons in further regions are activated in advance without receiving any information.
To fuse membrane potential smoothly, we design a project function $\Omega ( \cdot )$ to project the mixed membrane potential into an optimized space as Eq. \ref{eq4.1}.

\subsection{Theoretical Analysis}
The impacts of utilizing multi-direction membrane potential on SNNs are analyzed in this section. 
With theoretical tools related to the firing reset mechanism of LIF, we discover that our framework can significantly improve spike rates during the training process. Furthermore, the effect of project function $\Omega( \cdot)$ with different radian factors $\theta$ is also explained. \par
\textbf{Multi-direction membrane potential.}
Eq. \ref{eq3.1} to Eq. \ref{eq3.3} in the preliminary section demonstrate that conventional LIF neurons receive a spiking signal $X_l(t)$ and accumulate membrane potential $M_l(t)$ from the previous moment $t-1$, where $t$ is an element of the set ${(1, 2, ..., T)}$. However, it is expensive to calculate neuron states by directly solving a continuous function, the neuronal dynamics can be generally done by discretely evaluating the equations over small time steps as:
\begin{equation}
V[ n] =\alpha V[ n-1] +\ I[n]\ -\delta o[n-1]\ \ \ \ \
o[n] = 
\begin{cases}
    1 & \text{ if } V[n]>\delta \\
0  & \text{ otherwise }
\end{cases}
\label{eq4.3.5}
\end{equation}
where $n$ represents an discrete index of simulation time step, $\alpha=-\tau _{mem}$ that a leakage coefficient. When the neurons $w_{j}$ align with the preliminary findings, Eq. \ref{eq4.3.6} and Eq. \ref{eq4.3.7} can be utilized to show a contrast with Eq. \ref{eq3.1} and Eq. \ref{eq3.2}. In particular, the term $I(l,\mathbf{t})$ denotes the synaptic current, which is a differential function that relates to layer $l$ and time steps $t$ (assuming that the time step is limited to 1, we treat $t$ as a constant hereafter). And the term $V_{p}(G,\mathbf{t})$ represents the membrane potential, which is a differential function that depends on regions of multi-pole direction $G$ and not merely on the inputs at each time steps $t$.  \par
\begin{equation}
\frac{\partial I(l,\mathbf{t})}{\partial l} = -\tau _{syn}\lim\limits _{\mathbf{t}\rightarrow 1} I(l,\mathbf{t})+\sum _{j=1}^{J} (w_{j} ,X_{j} (l,\mathbf{t}))
\label{eq4.3.6}
\end{equation}

\begin{equation}
\frac{\partial V_{p} (G,\mathbf{t})}{\partial G} =-\tau _{mem}\lim\limits _{\mathbf{t}\rightarrow 1} V_{p} (G,\mathbf{t})+\ I (l,\mathbf{t})\ -\delta o_{l-1}^{\mathbf{t}}
\label{eq4.3.7}
\end{equation}

To demonstrate the superiority of the proposed framework explicitly, we employ the Taylor formula to approximate the truth by discretizing the continuous values $\partial I(t)$ and $\partial V(t)$, as shown in Eq. \ref{eq4.3.8}. This discretization preserves the same form as the common LIF model.
The detailed mathematical derivation is shown in Appendix.
\begin{equation}
V[\eta ,\iota ,n]=\alpha V_{p} [\eta -1,\iota -1,n]+\ I_{p} [\iota ,n]\ -\delta o[\iota -1]\ \ ( n=1)
\label{eq4.3.8}
\end{equation}
where $\eta$ and $\iota$ represent the unit separation of two different potential directions, $\alpha$ is an estimated leakage coefficient after fusion.

The synaptic currents of the spiking neurons are multiplied by the same constant at every time step. This leads to fast-diminishing gradients during a backward pass, as shown in previous research \cite{ponghiran2022spiking}. In our work, updating the synaptic currents is solely determined by the inputs at each layer, regardless of their existing values. This approach avoids a large number of extra calculations and latency ($\times T$), and compensates for the inadequacy-activation of neurons that accumulate membrane potential with limited time steps $t$. Neurons in each region accumulate membrane potential in both directions and the closer a region approaches $P$, the more times it will accumulate.
\begin{equation}
N_{convention}^{Accumulation}[m_{G_p}*T]\approx N_{bi-direction}^{Accumulation}[(m_{G_1}+m_{G_2}+...+m_{G_p})*1]
\label{eq4.3.9}
\end{equation}
Herein, as $p$ increases, the number of times that the membrane potential accumulates approximately equals that obtained using a large number of T. \par 
\textbf{Project function $\Omega( \cdot)$ and radian factors $\theta$.} 
Function $\Omega( \cdot)$ projects membrane potential onto an optimal fusion surface. Different radian factors $\theta$ determine the amplitude and steepness of the surface. To avoid the membrane potential tending towards negative infinity or over-activation (exceeding the threshold multiple times) after fusion, we select the surface by adjusting $\theta$. Regardless of $\theta$, the negative infinity of the fused membrane potential is resolved as mentioned above. Besides, to detailed analysis over-activation issue, we use the area difference $\nabla D$ ranged in $R \in (th,2*th )$ between the fusion surface ($\mathbb{S}_{Surface}=\Omega(\mathbb{X},\mathbb{Y})$) and the linear plane ($\mathbb{S}_{Plane}=\mathbb{X}+\mathbb{Y}$) as an indicator. This range is chosen because linear fusion only causes over-activation within this range. The area difference can be represented by Eq. \ref{eq.area_difference}. Note that since both the surface and the plane are origin-symmetric, the formula only applies to two positive shaft directions ($x$ and $y$).
\begin{equation}
    \nabla D=\mathbb{S}_{Surface} -\mathbb{S}_{Plane} 
\end{equation}

\begin{gather}
\mathbb{S}_{Surface} =\int\nolimits _{th}^{2*th}\int\nolimits _{th}^{2*th} \si{\ohm}( M_{1} ,M_{2}|\theta) dm_{1} dm_{2} =\mathbb{S}(\theta)\notag\\
\mathbb{S}_{Plane} =\int\nolimits _{th}^{2*th}\int\nolimits _{th}^{2*th}( M_{1} +M_{2}) dm_{1} dm_{2} =3( th)^{3}
\label{eq.area_difference}
\end{gather}

The area of the projection plane $\mathbb{S}_{Plane}$ is determined by the threshold of the LIF model, the difference $\nabla D$ depends only on the radian factors $\theta$. Theoretically, the greater the absolute difference, the greater the possibility of over-activation. 

Results and detailed calculation process of the difference $\nabla D$ are presented in the supplementary material (consistent with the experiment, $th$ is set as 0.5). The over-activation of spiking neurons caused by membrane potential fusion can be suppressed with each $\theta$.
More specifically, according to the experimental results in section 4.4, CNN and RNN-based models shows different association with $\theta$.

\begin{table}[t]
\caption{\centering{Performance on Cifar10, Cifar100, and Tiny-ImageNet datasets.}}
\label{table.4.1}
\centering
\begin{threeparttable}
\resizebox{\columnwidth}{!}{%
\begin{tabular}{cllllcc}
\toprule
\textbf{Datasets} & \multicolumn{2}{c}{\textbf{Method}} & \textbf{Type} & \textbf{Baseline} & \textbf{Acc(\%)} & \textbf{Time steps(T)} \\ \midrule
\multirow{13}{*}{\textbf{Cifar10}} & \multicolumn{2}{l}{SDN-LIF\cite{hunsberger2015spiking}} & ANN Conversion & 2C, 2L & 82.95 & 6000 \\
 & \multicolumn{2}{l}{IOM\cite{bu2022optimized}} & ANN Conversion & ResNet-20 & 92.75 & 512 \\
 & \multicolumn{2}{l}{SpikeConverter\cite{liu2022spikeconverter}} & ANN Conversion & ResNet-20 & 91.47 & 16 \\
 & \multicolumn{2}{l}{Diet-SNN\cite{rathi2021diet}} & C\&T\tnote{*} & ResNet-20 & 92.54 & 10 \\
 & \multicolumn{2}{l}{AutoSNN\cite{na2022autosnn}} & SNN Training & NAS64 & 92.54 & 8 \\
 & \multicolumn{2}{l}{STBP-tdBN\cite{zheng2021going}} & SNN Training & ResNet-19 & 93.16 & 6 \\
 & \multicolumn{2}{l}{MLF(K=1)\cite{feng2022multi}} & SNN Training & DS-ResNet-12 & 92.46 & 4 \\
 & \multicolumn{2}{l}{Dspike SNN\cite{li2021differentiable}} & SNN Training & ResNet-18 & 93.13 & 2 \\
 & \multicolumn{2}{l}{Emporal pruning\cite{chowdhury2022towards}} & SNN Training & VGG16 & 93.05 & 5-1 \\ \cline{2-7} 
 & \multirow{4}{*}{\textbf{\begin{tabular}[c]{@{}l@{}}This\\ work\end{tabular}}} & CNN & \multirow{4}{*}{SNN Training} & ResNET-12F & \textbf{93.07} & \textbf{1} \\
 &  & LSTM &  & VIT-12 & 92.68 & \textbf{1} \\
 &  & CNN-MSp\tnote{**} &  & ResNET-12F & \textbf{93.27} & \textbf{1} \\
 &  & LSTM-MSp &  & VIT-12 & 93.13 & \textbf{1} \\ \hline
\multirow{8}{*}{\textbf{Cifar100}} & \multicolumn{2}{l}{RMP-SNN\cite{han2020rmp}} & ANN Conversion & VGG16 & 70.93 & 2048 \\
 & \multicolumn{2}{l}{IOM\cite{bu2022optimized}} & ANN Conversion & ResNet-20 & 70.51 & 128 \\
 & \multicolumn{2}{l}{Diet-SNN\cite{rathi2021diet}} & C\&T & ResNet-20 & 69.67 & 5 \\
 & \multicolumn{2}{l}{AutoSNN\cite{na2022autosnn}} & SNN Training & NAS64 & 69.16 & 8 \\
 & \multicolumn{2}{l}{Dspike SNN\cite{li2021differentiable}} & SNN Training & ResNet-18 & 71.68 & 2 \\
 & \multicolumn{2}{l}{Emporal pruning\cite{chowdhury2022towards}} & SNN Training & VGG16 & 70.15 & 5-1 \\ \cline{2-7} 
 & \multirow{2}{*}{\textbf{\begin{tabular}[c]{@{}l@{}}This\\ work\end{tabular}}} & CNN & \multirow{2}{*}{SNN Training} & ResNET-12F & \textbf{72.41} & \textbf{1} \\
 &  & LSTM &  & VIT-12 & \textbf{72.31} & \textbf{1} \\ \hline
\multicolumn{1}{l}{\multirow{6}{*}{\begin{tabular}[c]{@{}l@{}}\textbf{Tiny}\\ \textbf{ImageNet}\end{tabular}}} & \multicolumn{2}{l}{Spike-Norm\cite{na2022autosnn}} & ANN Conversion & VGG16 & 48.60 & 2500 \\
\multicolumn{1}{l}{} & \multicolumn{2}{l}{Spike-Thrift\cite{kundu2021spike}} & C\&T & VGG16 & 51.92 & 150 \\
\multicolumn{1}{l}{} & \multicolumn{2}{l}{CAT\cite{lew2022time}} & SNN Training & VGG16 & 57.40 & 48 \\
\multicolumn{1}{l}{} & \multicolumn{2}{l}{AutoSNN\cite{na2022autosnn}} & SNN Training & NAS64 & 46.79 & 8 \\ \cline{2-7} 
\multicolumn{1}{l}{} & \multirow{2}{*}{\textbf{\begin{tabular}[c]{@{}l@{}}This\\ work\end{tabular}}} & CNN & SNN Training & ResNET-12F & \textbf{57.91} & \textbf{1} \\
\multicolumn{1}{l}{} &  & LSTM & SNN Training & VIT-12 & \textbf{57.55} & \textbf{1} \\ \bottomrule
\end{tabular}
}
    \begin{tablenotes}
        \footnotesize
        \item[*] A combination method of ANN conversion and direct training
        \item[**] The case where Median spike (Ms) is not activated to a complete spike.
      \end{tablenotes}
      \end{threeparttable}
\label{table.e5.1}
\end{table}
\section{Experiments and Results}
Please refer to the supplementary material for details on the experimental settings. \par

\subsection{Performance Comparison}

We compare our models (CNN/LSTM) with various state-of-the-art (SOTA) SNNs, and the results are shown in Table \ref{table.4.1}. To adapt different models, we choose ResNet-12 \cite{touvron2022resmlp} as the baseline network for the CNN model, we have show the reason in section 3. Besides, VIT-12 is for the RNN model (the networtk of residual series is not suitable for RNN). Since LSTM outperforms other RNN-based methods, we only report the results of LSTM-based models. The results of other RNN-based models are included in the Appendix. Our CNN-based models achieve top-1 accuracy of 93.07\%, 72.41\%, and 57.91\% on CIFAR-10, CIFAR-100, and Tiny-ImageNet, respectively, with just 1 time step (T=1). Analogously, the top-1 accuracies for the LSTM-based models are 93.07\%, 72.31\%, and 57.55\% regarding the same datasets.
In terms of efficiency, SNN performs equally well or better than other models in terms of accuracy with our framework, while achieving significantly lower inference latency. Importantly, proposal enables us to reduce the SNN latency to the lowest possible limit (\textbf{time steps = 1}) without the need for pre-training. Furthermore, to our knowledge, the RNN-SNN method is the first to achieve comparable results to CNN-based models for visual classification tasks.

\subsection{Energy Consumption}

\begin{table*}[t!]
\centering
\newcolumntype{C}[1]{>{\centering\arraybackslash}p{#1}}
\caption{Computing cost and energy consumption in 45nm CMOS on Cifar10.}
\begin{tabular}{C{3cm}C{4cm}C{1cm}C{1cm}C{1cm}}
\toprule 
 Method & Model & \#Add & \#Mult & Energy \\
\midrule 
 ANN & Res-19 & 2285M & 2285M & 12.6$J$ \\
ANN(ours) & ResNET-12F & 247M & 247M & 1.35$J$ \\
ANN(ours) & VIT-12 & 251M & 251M & 1.38$J$ \\
\hline 
 SNN & Res-20(T=5) & 142M & 8.80M & 168$mJ$ \\
SNN [26] & Res-19(T=2) & 360M  & 6.80M  & 355$mJ$ \\
SNN(ours) & Res12F-CNN(T=1) & 132M & 2.14M & 128$mJ$ \\
SNN(ours) & VIT-12-LSTM(T=1) & 79M & 5.31M & 96$mJ$ \\
 \bottomrule
\end{tabular}
 \label{table5.2}
\end{table*}

In ANN, each operation computes a dot product involving one multiplication and addition (MAC) in floating-point (FP) format, whereas, in SNNs, the multiplication is eliminated by the binary spike. 
Energy is therefore saved due to a large number of more expensive multiplication has been replaced by addition operations.
Namely, SNN demonstrates the energy efficiency gains by the cheaper multiplexer and compactor due to the event-driven paradigm \cite{horowitz20141}.  
In this manner, the addition operation is a primary energy consumer in SNNs, and its frequency is largely determined by the spike rate.  
The spike rate in iso-architecture SNN is commonly specified as Eq. \ref{eq.spikerate}. 
\begin{equation}
\#Spiking-Rate_{l} =\frac{\#TotalSpike_{l} \times Timesteps}{\#Neurons_{l}}
\label{eq.spikerate}
\end{equation}
where $\#Spike-Rate$ is the total spikes in layer $l$ over all times steps averaged over the number of neurons in layer $l$. Spike rate may exceed 1 in some studies (by over numerous time steps) implying that the number of operations for SNN exceeds the ANN (operations are MAC in ANN while still adding in SNNs). 
Explicitly, lower spike rates denote more sparsity in spike events, lower operations, and higher energy efficiency for SNN. \par
SNNs training form our method, the average spike rate is 0.54 and 0.32 for CNN-SNN and LSTM-SNN, respectively. Following previous work \cite{horowitz20141}, we estimate the energy consumption of SNNs by comparing the energy consumption of SNN and ANN in 45 nm CMOS technology. The energy cost for a 32-bit ANN MAC operation (4.6 pJ) is 5.1 $\times$ higher than that of an SNN addition operation (0.9 pJ). The number of operations or layers in an ANN is defined by Eq. \ref{eq.ANNOP}.
\begin{equation}
{\textstyle \#OP_{ANN} =\left\{\begin{array}{ l l }
 \begin{array}{l}
k_{w} \times k_{h} \times c_{in} \times h_{out} \times w_{out} \times c_{out}\\
( c_{in} +c_{out}) \times c_{out} \times 4\ \\
f_{in} \times f_{out}
\end{array}  &  \begin{array}{l}
,Convolution\ layer\\
,LSTM\ layer\\
,Fully-connected\ layer
\end{array}
\end{array}\right.} 
\label{eq.ANNOP}
\end{equation}
where $k_{w}(k_{h})$ is kernel width (height), $c_{in}(c_{out})$ is the number of input (output) channels, $h_{out}(w_{out})$ is the height (width) of the output feature map, and $f_{in}(f_{out})$ is the channels of input (output) features.

\begin{figure}[t!]
\centerline{\includegraphics[width=0.8\linewidth]{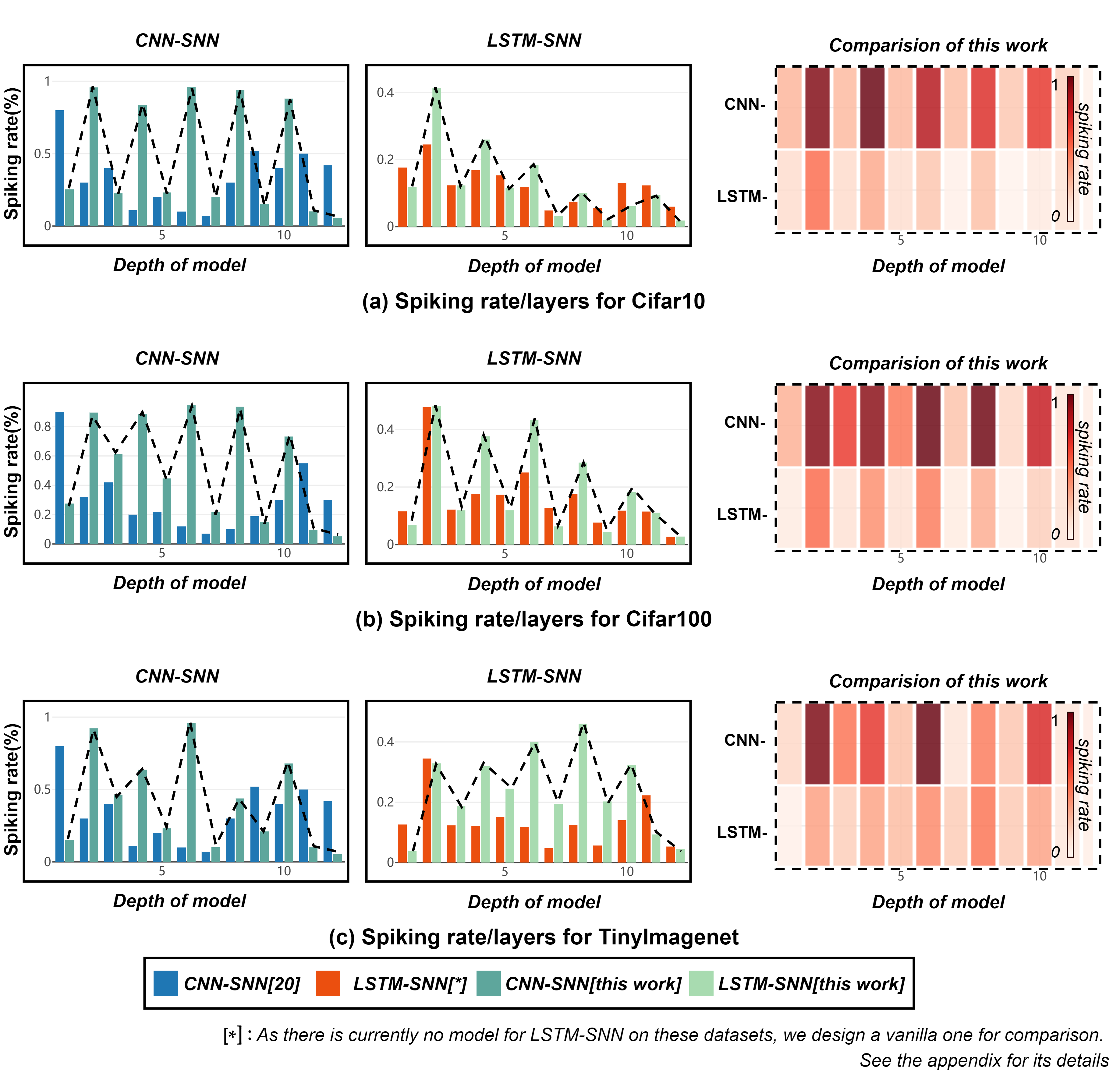}}
\caption{Comparison for spiking rate/layers based on various frameworks.}
\label{Heat_map}
\end{figure}

For an SNN, the number of operations per layer is given by $\#OP_{SNN,l} =\#Spiking-Rate_{l} \ \times \ {\textstyle \#OP_{ANN}}$, where $\#OP_{SNN,l}$ represents the total number of MAC operations in layer $l$. Using these equations, the addition count is calculated by $s \ *\ T \ *\ \#OP_{ANN}$ in SNN, where $s$ is the mean spike rate, and $T$ is the time steps. For multiplication in the SNN, we set it to the MAC of the first layer and some fully-connected layers in our model and scale it by $T$. We calculate the operation number and energy consumption for both the ANN and SNN, which is shown in Table \ref{table5.2}. Our model only costs 123 mJ for a single forward pass, which is 11 $\times$ lower in energy consumption than our ANN, and 102 $\times$ lower than Res-19 in the work \cite{zheng2021going}. 
As shown in Fig. \ref{Heat_map}, the proposed framework enables spike rate changes alternately between different layers. Leaving out the last block, the spike rate will be greatly enhanced in even layers (dilated window) and remain low in odd layers (non-dilated window). However, as the network is continuously halved, the feature maps in the last layers generally become small, and the dilated window may not be appropriately divided (such as the size is $2 \times 2$). Therefore, we adopt a non-dilated window in the last block of the framework.

\begin{table*}[t!]
  \centering
  \caption{Spike/image comparison between this work and SOTA SNN \cite{tang2022snn2ann}.}
  \renewcommand{\arraystretch}{1.2}
  \resizebox{\columnwidth}{!}{%
    \begin{tabular}{|c|c|c|c|c|c|c|c|c|c|}
    \hline
    \multirow{2}[4]{*}{\textbf{Method}} & \multicolumn{3}{c|}{\textbf{CIFAR10}} & \multicolumn{3}{c|}{\textbf{CIFAR100}} & \multicolumn{3}{c|}{\textbf{Tiny-ImageNet}} \\
\cline{2-10}          & \textbf{Acc (\%)} & \textbf{Steps} & \textbf{Spikes/Image} & \textbf{Acc (\%)} & \textbf{Steps} & \textbf{Spikes/Image} & \textbf{Acc (\%)} & \textbf{Steps} & \textbf{Spikes/Image} \\
    \hline
    RNL-RIL & 92.50  & 250   & $4.24\times 10^6$ & 72.90 & 250   & $5.94\times 10^6$ & 56.10 & 250   & $2.03\times 10^7$ \\
    \hline
    \multirow{2}{*}{STBP-IF} & 84.20  & 5     & $1.11\times 10^6$ & 57.77  & 4     & $1.35\times 10^6$ & \multirow{2}{*}{54.53} & \multirow{2}{*}{3} & \multirow{2}{*}{$2.17\times 10^6$} \\
\cline{2-7}          & 71.54  & 8     & $2.01\times 10^6$ & 39.86  & 8     & $2.69\times 10^6$ &       &       &  \\
    \hline
    STBP-PLIF & 91.63  & 5     & $9.67\times 10^5$ & 70.94  & 4     & $1.05\times 10^6$ & 53.08  & 3     & $1.74\times 10^6$ \\
    \hline
    S2A-ReSU & 92.62 & 5     & $1.68\times 10^6$ & 71.10 & 4     & $6.69\times 10^5$ & 54.91 & 3     & $1.02\times 10^6$ \\
    \hline
    S2A-STSU & 92.18 & 5     & $4.52\times 10^5$ & 68.96  & 4     & $6.18\times 10^5$ & 54.33  & 3     & $1.11\times 10^6$ \\
    \hline
    \textbf{Ours-CNN} & 93.07 & 1     & $\mathbf{6.38\times 10^4}$ & 72.41  & 1     & $\mathbf{6.19\times 10^4}$ & 57.91  & 1     & $\mathbf{2.58\times 10^5}$ \\
    \hline
    \textbf{Ours-LSTM} & 92.68 & 1     & $\mathbf{2.13\times 10^4}$ & 72.31  & 1     & $\mathbf{2.44\times 10^4}$ & 57.55  & 1     & $\mathbf{1.85\times 10^5}$ \\
    \hline
    
    \end{tabular}%
    }
    
  \label{tab:same_cmp}%
\end{table*}%

From a spike rate perspective, we use a higher average spike rate than other methods. However, this higher spike rate is compensated for by our model's lightweight and low-latency nature. When noticing the total amount of spike activity, as 12 network layers and only one-time step in our model, the total spike activity remains on a small scale, as shown in Table \ref{tab:same_cmp}. In most cases, our SNNs achieve comparable accuracy with fewer spikes than other methods. 


\subsection{Ablations}
We present an ablation study of our proposal, divided into two aspects corresponding to Sections \ref{section3.1} and \ref{section3.2}. First, we ablate the substructure dilated windows of the proposed framework. Second, we experiment with other fusion functions and compare them to our proposed fusion function using different values of $\theta$. \par
\textbf{Dilated Window} Ablations of the dilated window approach on Cifar10 for both frameworks are reported in Table \ref{ab.1}. We first analyze the effect of windowing times (i.e., the size of $i$ in $G_m^i$) on the CNN-based model without dilated window. Next, we verify the dilated window of different windowing times. Finally, we do the same experiment with the LSTM-SNN model. \par
From the results, we observe that reducing the number of windowing partitions leads to a decrease in performance, as we explained in Section \ref{section3.1}. The multiple uses of membrane potential are positively correlated with the performance of SNN within a limited number of time steps. Additionally, we verify the effectiveness of our proposed Dilated Window approach. Due to the strengthened connection between different windows, its performance is better than that of the normal Window partition method.

\begin{table}[t!]
    \caption{Ablation study of proposed window partition.}
    \centering
    \resizebox{\columnwidth}{!}{%
        \begin{tabular}{c|c|c|c|c|c|c|cc|c}
        \toprule
         \multicolumn{5}{c|}{CNN-SNN} & \multicolumn{5}{c}{LSTM-SNN} \\
        \hline 
         Groups & g=1 & g=2 & g=4 & Acc\%) & Groups & g=1 & \multicolumn{1}{c|}{g=2} & \multicolumn{1}{c|}{g=4} & Acc(\%) \\
        \hline 
         \multirow{3}{*}{Normal Windows} & \checkmark &  &  & 91.80 & \multirow{3}{*}{Normal Windows} & \checkmark & \multicolumn{1}{c|}{} & \multicolumn{1}{c|}{} & 90.37 \\
        \cline{2-5} \cline{7-10} 
           &  & \checkmark &  & 92.15 &   &  & \multicolumn{1}{c|}{\checkmark} & \multicolumn{1}{c|}{} & 90.02 \\
        \cline{2-5} \cline{7-10} 
           &  &  & \checkmark & 92.28 &   &  & \multicolumn{1}{c|}{} & \checkmark & 90.58 \\
        \hline 
         \multirow{3}{*}{Dilated Windows} & \checkmark &  &  & 91.80 & \multirow{3}{*}{Dilated Windows} & \checkmark & \multicolumn{1}{c|}{} & \multicolumn{1}{c|}{} & 90.37 \\
        \cline{2-5} \cline{7-10} 
           &  & \checkmark &  & 92.78 &   &  & \multicolumn{1}{c|}{\checkmark} & \multicolumn{1}{c|}{} & 91.82 \\
        \cline{2-5} \cline{7-10} 
           &  &  & \checkmark & \textbf{93.13} &   &  & \multicolumn{1}{c|}{} & \multicolumn{1}{c|}{\checkmark} & \textbf{92.58} \\
         \bottomrule
        \end{tabular}   
        }
    \label{ab.1}
\end{table}

\begin{figure}[t!]
\centerline{\includegraphics[width=0.8\linewidth]{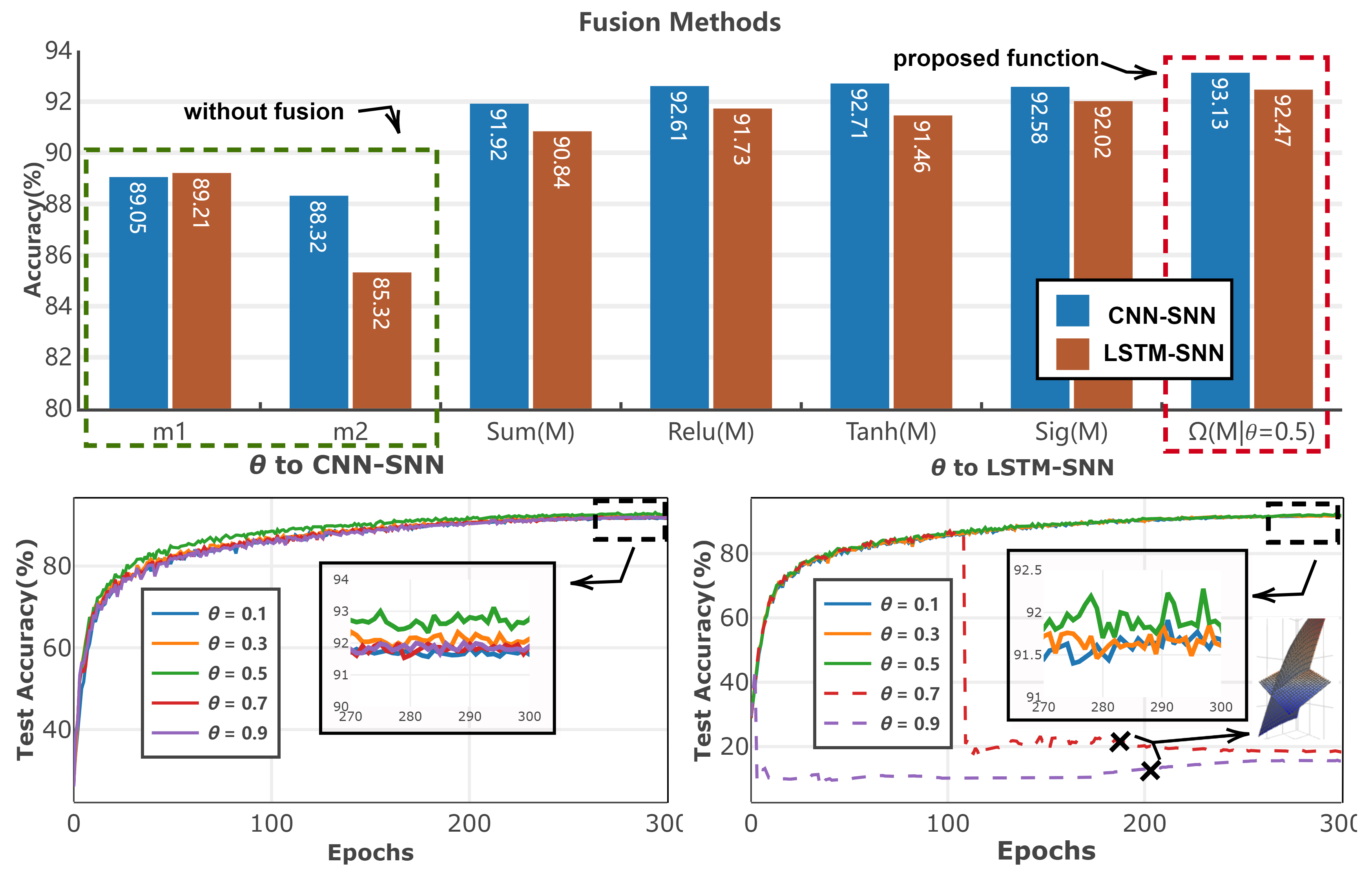}}
\caption{Ablation study of project functions and various $\theta$.}
\label{theta}
\end{figure}
\par
\textbf{Projection Function $\Theta ( \cdot )$} 
To verify its effectiveness, the projection function is ablated in two aspects, as shown in Fig. \ref{theta}. First, several mainstream nonlinear functions (Relu, Sigmoid, Tanh) are experimented with and compared to the proposed functions to explore an optimal fusion space for membrane potential. Second, the impact of different angles $\theta$ on the proposed projection function is investigated. 

As shown in Fig. \ref{theta}(a), the proposed function performs best under insufficient time step constraints. Creating a remarkable SNN through only one direction of membrane potential is challenging; other nonlinear functions have not yielded better results than the proposed function. This is because nonlinear functions avoid negative infinity of membrane potential, resulting in better performance than the linear function. Additionally, while over-activation can only be prevented from worsening in other functions, the symmetry of the trigonometric function allows neurons to inhibit membrane potential fusion within a certain range. In contrast, as shown in Fig. \ref{theta}(b)(c), the model achieves the best convergence effect when $\theta=0.5$. Varying $\theta$ will affect the final convergence of the model, as previously analyzed. However, it should be noted that the convergence of RNN-SNN is relatively unstable and may even fail, as indicated by the dotted lines. This is due to the rapid attenuation of membrane potential after fusion caused by excessive $\theta$ as illustrated in the auxiliary spatial map next to the dotted line. It should be noted that the threshold selection of LIF is critical for fusion. We chose $th=0.5$ because it can make the double threshold not greater than the maximum value of the trigonometric function. 

\section{Conclusion $\&$ Discussion}
In this paper, we propose a novel training framework for Spiking Neural Networks, which results in low inference latency from $T$ time steps down to unity ($1$) while maintaining competitive performance. Through enhancing the utilization of membrane potential information in two directions, the learning ability of SNNs has been significantly improved in an extremely limited time step as shown in both experimental results and theoretical analysis. Furthermore, the proposed framework is also applicable to RNN-based models, achieving unprecedented results across multiple datasets. Although our model demonstrates significant performance improvement, it does not result in a more sparse activation spike matrix when compared to previous models. The advantages in energy consumption mainly arise from the extremely minimal time steps used in our work. Developing a more sparse model to directly deal with the inherent information loss in discrete spikes remains a problem. Additionally, the design of the projection function and $\theta$ is based on previous research and experience on spiking neurons, and may not be optimal. Other types of fusion methods for membrane potential are worth exploring further.

\acks{This research was supported by the JSPS KAKENHI Grant Number JP22K21285, Japan.}

\bibliography{acml23}

\begin{thebibliography}{49}
\providecommand{\natexlab}[1]{#1}
\providecommand{\url}[1]{\texttt{#1}}
\expandafter\ifx\csname urlstyle\endcsname\relax
  \providecommand{\doi}[1]{doi: #1}\else
  \providecommand{\doi}{doi: \begingroup \urlstyle{rm}\Url}\fi

\bibitem[Bellec et~al.(2018)Bellec, Salaj, Subramoney, Legenstein, and Maass]{bellec2018long}
Guillaume Bellec, Darjan Salaj, Anand Subramoney, Robert Legenstein, and Wolfgang Maass.
\newblock Long short-term memory and learning-to-learn in networks of spiking neurons.
\newblock \emph{Advances in neural information processing systems}, 31, 2018.

\bibitem[Bengio et~al.(2013)Bengio, L{\'e}onard, and Courville]{bengio2013estimating}
Yoshua Bengio, Nicholas L{\'e}onard, and Aaron Courville.
\newblock Estimating or propagating gradients through stochastic neurons for conditional computation.
\newblock \emph{arXiv preprint arXiv:1308.3432}, 2013.

\bibitem[Bu et~al.(2022)Bu, Ding, Yu, and Huang]{bu2022optimized}
Tong Bu, Jianhao Ding, Zhaofei Yu, and Tiejun Huang.
\newblock Optimized potential initialization for low-latency spiking neural networks.
\newblock In \emph{Proceedings of the AAAI Conference on Artificial Intelligence}, volume~36, pages 11--20, 2022.

\bibitem[Cao et~al.(2015)Cao, Chen, and Khosla]{cao2015spiking}
Yongqiang Cao, Yang Chen, and Deepak Khosla.
\newblock Spiking deep convolutional neural networks for energy-efficient object recognition.
\newblock \emph{International Journal of Computer Vision}, 113\penalty0 (1):\penalty0 54--66, 2015.

\bibitem[Chen et~al.(2022)Chen, Zhu, Yang, and Zhang]{ChenIJCAI}
Zheng Chen, Lingwei Zhu, Ziwei Yang, and Renyuan Zhang.
\newblock Multi-tier platform for cognizing massive electroencephalogram.
\newblock In \emph{Proceedings of the Thirty-First International Joint Conference on Artificial Intelligence, {IJCAI}}, pages 2464--2470, 2022.

\bibitem[Chowdhury et~al.(2022{\natexlab{a}})Chowdhury, Rathi, and Roy]{ECCV-2022-roy}
Sayeed~Shafayet Chowdhury, Nitin Rathi, and Kaushik Roy.
\newblock Towards ultra low latency spiking neural networks for vision and sequential tasks using temporal pruning.
\newblock In \emph{ECCV}, pages 709--726, 2022{\natexlab{a}}.

\bibitem[Chowdhury et~al.(2022{\natexlab{b}})Chowdhury, Rathi, and Roy]{chowdhury2022towards}
Sayeed~Shafayet Chowdhury, Nitin Rathi, and Kaushik Roy.
\newblock Towards ultra low latency spiking neural networks for vision and sequential tasks using temporal pruning.
\newblock In \emph{Computer Vision--ECCV 2022: 17th European Conference, Tel Aviv, Israel, October 23--27, 2022, Proceedings, Part XI}, pages 709--726. Springer, 2022{\natexlab{b}}.

\bibitem[Datta et~al.(2022{\natexlab{a}})Datta, Deng, Aviles, and Beerel]{AAAI-2023-Peter}
Gourav Datta, Haoqin Deng, Robert Aviles, and Peter~A. Beerel.
\newblock Towards energy-efficient, low-latency and accurate spiking lstms, 2022{\natexlab{a}}.

\bibitem[Datta et~al.(2022{\natexlab{b}})Datta, Deng, Aviles, and Beerel]{datta2022towards}
Gourav Datta, Haoqin Deng, Robert Aviles, and Peter~A Beerel.
\newblock Towards energy-efficient, low-latency and accurate spiking lstms.
\newblock \emph{arXiv preprint arXiv:2210.12613}, 2022{\natexlab{b}}.

\bibitem[Deng and Gu(2021)]{deng2021optimal}
Shikuang Deng and Shi Gu.
\newblock Optimal conversion of conventional artificial neural networks to spiking neural networks.
\newblock \emph{arXiv preprint arXiv:2103.00476}, 2021.

\bibitem[Dosovitskiy et~al.(2020)Dosovitskiy, Beyer, Kolesnikov, Weissenborn, Zhai, Unterthiner, Dehghani, Minderer, Heigold, Gelly, et~al.]{dosovitskiy2020image}
Alexey Dosovitskiy, Lucas Beyer, Alexander Kolesnikov, Dirk Weissenborn, Xiaohua Zhai, Thomas Unterthiner, Mostafa Dehghani, Matthias Minderer, Georg Heigold, Sylvain Gelly, et~al.
\newblock An image is worth 16x16 words: Transformers for image recognition at scale.
\newblock \emph{arXiv preprint arXiv:2010.11929}, 2020.

\bibitem[Fang et~al.(2021{\natexlab{a}})Fang, Yu, Chen, Huang, Masquelier, and Tian]{fang2021deep}
Wei Fang, Zhaofei Yu, Yanqi Chen, Tiejun Huang, Timoth{\'e}e Masquelier, and Yonghong Tian.
\newblock Deep residual learning in spiking neural networks.
\newblock \emph{Advances in Neural Information Processing Systems}, 34:\penalty0 21056--21069, 2021{\natexlab{a}}.

\bibitem[Fang et~al.(2021{\natexlab{b}})Fang, Yu, Chen, Masquelier, Huang, and Tian]{fang2021incorporating}
Wei Fang, Zhaofei Yu, Yanqi Chen, Timoth{\'e}e Masquelier, Tiejun Huang, and Yonghong Tian.
\newblock Incorporating learnable membrane time constant to enhance learning of spiking neural networks.
\newblock In \emph{Proceedings of the IEEE/CVF International Conference on Computer Vision}, pages 2661--2671, 2021{\natexlab{b}}.

\bibitem[Feng et~al.(2022)Feng, Liu, Tang, Ma, and Pan]{feng2022multi}
Lang Feng, Qianhui Liu, Huajin Tang, De~Ma, and Gang Pan.
\newblock Multi-level firing with spiking ds-resnet: Enabling better and deeper directly-trained spiking neural networks.
\newblock \emph{arXiv preprint arXiv:2210.06386}, 2022.

\bibitem[Han et~al.(2020)Han, Srinivasan, and Roy]{han2020rmp}
Bing Han, Gopalakrishnan Srinivasan, and Kaushik Roy.
\newblock Rmp-snn: Residual membrane potential neuron for enabling deeper high-accuracy and low-latency spiking neural network.
\newblock In \emph{Proceedings of the IEEE/CVF conference on computer vision and pattern recognition}, pages 13558--13567, 2020.

\bibitem[Horowitz(2014)]{horowitz20141}
Mark Horowitz.
\newblock 1.1 computing's energy problem (and what we can do about it).
\newblock In \emph{2014 IEEE International Solid-State Circuits Conference Digest of Technical Papers (ISSCC)}, pages 10--14. IEEE, 2014.

\bibitem[Hunsberger and Eliasmith(2015)]{hunsberger2015spiking}
Eric Hunsberger and Chris Eliasmith.
\newblock Spiking deep networks with lif neurons.
\newblock \emph{arXiv preprint arXiv:1510.08829}, 2015.

\bibitem[Kim et~al.(2018)Kim, Kim, Huh, Lee, and Choi]{kim2018deep}
Jaehyun Kim, Heesu Kim, Subin Huh, Jinho Lee, and Kiyoung Choi.
\newblock Deep neural networks with weighted spikes.
\newblock \emph{Neurocomputing}, 311:\penalty0 373--386, 2018.

\bibitem[Kiselev(2016)]{kiselev2016rate}
Mikhail Kiselev.
\newblock Rate coding vs. temporal coding-is optimum between?
\newblock In \emph{2016 international joint conference on neural networks (IJCNN)}, pages 1355--1359. IEEE, 2016.

\bibitem[Krizhevsky et~al.(2009)Krizhevsky, Hinton, et~al.]{krizhevsky2009learning}
Alex Krizhevsky, Geoffrey Hinton, et~al.
\newblock Learning multiple layers of features from tiny images.
\newblock 2009.

\bibitem[Kundu et~al.(2021)Kundu, Datta, Pedram, and Beerel]{kundu2021spike}
Souvik Kundu, Gourav Datta, Massoud Pedram, and Peter~A Beerel.
\newblock Spike-thrift: Towards energy-efficient deep spiking neural networks by limiting spiking activity via attention-guided compression.
\newblock In \emph{Proceedings of the IEEE/CVF Winter Conference on Applications of Computer Vision}, pages 3953--3962, 2021.

\bibitem[Le and Yang(2015)]{le2015tiny}
Ya~Le and Xuan Yang.
\newblock Tiny imagenet visual recognition challenge.
\newblock \emph{CS 231N}, 7\penalty0 (7):\penalty0 3, 2015.

\bibitem[Lew et~al.(2022)Lew, Lee, and Park]{lew2022time}
Dongwoo Lew, Kyungchul Lee, and Jongsun Park.
\newblock A time-to-first-spike coding and conversion aware training for energy-efficient deep spiking neural network processor design.
\newblock In \emph{Proceedings of the 59th ACM/IEEE Design Automation Conference}, pages 265--270, 2022.

\bibitem[Li et~al.(2022)Li, Chen, Guo, Zhang, and Wang]{li2022brain}
Wenshuo Li, Hanting Chen, Jianyuan Guo, Ziyang Zhang, and Yunhe Wang.
\newblock Brain-inspired multilayer perceptron with spiking neurons.
\newblock In \emph{Proceedings of the IEEE/CVF Conference on Computer Vision and Pattern Recognition}, pages 783--793, 2022.

\bibitem[Li et~al.(2021{\natexlab{a}})Li, Deng, Dong, Gong, and Gu]{li2021free}
Yuhang Li, Shikuang Deng, Xin Dong, Ruihao Gong, and Shi Gu.
\newblock A free lunch from ann: Towards efficient, accurate spiking neural networks calibration.
\newblock In \emph{International Conference on Machine Learning}, pages 6316--6325. PMLR, 2021{\natexlab{a}}.

\bibitem[Li et~al.(2021{\natexlab{b}})Li, Guo, Zhang, Deng, Hai, and Gu]{li2021differentiable}
Yuhang Li, Yufei Guo, Shanghang Zhang, Shikuang Deng, Yongqing Hai, and Shi Gu.
\newblock Differentiable spike: Rethinking gradient-descent for training spiking neural networks.
\newblock \emph{Advances in Neural Information Processing Systems}, 34:\penalty0 23426--23439, 2021{\natexlab{b}}.

\bibitem[Liu et~al.(2022)Liu, Zhao, Chen, Wang, and Jiang]{liu2022spikeconverter}
Fangxin Liu, Wenbo Zhao, Yongbiao Chen, Zongwu Wang, and Li~Jiang.
\newblock Spikeconverter: An efficient conversion framework zipping the gap between artificial neural networks and spiking neural networks.
\newblock In \emph{Proceedings of the AAAI Conference on Artificial Intelligence}, volume~36, pages 1692--1701, 2022.

\bibitem[Liu et~al.(2021)Liu, Lin, Cao, Hu, Wei, Zhang, Lin, and Guo]{liu2021swin}
Ze~Liu, Yutong Lin, Yue Cao, Han Hu, Yixuan Wei, Zheng Zhang, Stephen Lin, and Baining Guo.
\newblock Swin transformer: Hierarchical vision transformer using shifted windows.
\newblock In \emph{Proceedings of the IEEE/CVF International Conference on Computer Vision}, pages 10012--10022, 2021.

\bibitem[Lotfi~Rezaabad and Vishwanath(2020)]{lotfi2020long}
Ali Lotfi~Rezaabad and Sriram Vishwanath.
\newblock Long short-term memory spiking networks and their applications.
\newblock In \emph{International Conference on Neuromorphic Systems 2020}, pages 1--9, 2020.

\bibitem[Masquelier and Thorpe(2007)]{masquelier2007unsupervised}
Timoth{\'e}e Masquelier and Simon~J Thorpe.
\newblock Unsupervised learning of visual features through spike timing dependent plasticity.
\newblock \emph{PLoS computational biology}, 3\penalty0 (2):\penalty0 e31, 2007.

\bibitem[Na et~al.(2022)Na, Mok, Park, Lee, Choe, and Yoon]{na2022autosnn}
Byunggook Na, Jisoo Mok, Seongsik Park, Dongjin Lee, Hyeokjun Choe, and Sungroh Yoon.
\newblock Autosnn: towards energy-efficient spiking neural networks.
\newblock In \emph{International Conference on Machine Learning}, pages 16253--16269. PMLR, 2022.

\bibitem[Neftci et~al.(2019)Neftci, Mostafa, and Zenke]{neftci2019surrogate}
Emre~O Neftci, Hesham Mostafa, and Friedemann Zenke.
\newblock Surrogate gradient learning in spiking neural networks: Bringing the power of gradient-based optimization to spiking neural networks.
\newblock \emph{IEEE Signal Processing Magazine}, 36\penalty0 (6):\penalty0 51--63, 2019.

\bibitem[Ponghiran and Roy(2022)]{ponghiran2022spiking}
Wachirawit Ponghiran and Kaushik Roy.
\newblock Spiking neural networks with improved inherent recurrence dynamics for sequential learning.
\newblock In \emph{Proceedings of the AAAI Conference on Artificial Intelligence}, volume~36, pages 8001--8008, 2022.

\bibitem[Rathi and Roy(2020)]{rathi2020diet}
Nitin Rathi and Kaushik Roy.
\newblock Diet-snn: Direct input encoding with leakage and threshold optimization in deep spiking neural networks.
\newblock \emph{arXiv preprint arXiv:2008.03658}, 2020.

\bibitem[Rathi and Roy(2021{\natexlab{a}})]{TNN-2021-roy}
Nitin Rathi and Kaushik Roy.
\newblock Diet-snn: A low-latency spiking neural network with direct input encoding and leakage and threshold optimization.
\newblock \emph{IEEE Transactions on Neural Networks and Learning Systems}, pages 1--9, 2021{\natexlab{a}}.

\bibitem[Rathi and Roy(2021{\natexlab{b}})]{rathi2021diet}
Nitin Rathi and Kaushik Roy.
\newblock Diet-snn: A low-latency spiking neural network with direct input encoding and leakage and threshold optimization.
\newblock \emph{IEEE Transactions on Neural Networks and Learning Systems}, 2021{\natexlab{b}}.

\bibitem[Rathi et~al.(2020)Rathi, Srinivasan, Panda, and Roy]{ICLR-2022-Rathi}
Nitin Rathi, Gopalakrishnan Srinivasan, Priyadarshini Panda, and Kaushik Roy.
\newblock Enabling deep spiking neural networks with hybrid conversion and spike timing dependent backpropagation.
\newblock In \emph{International Conference on Learning Representations}, 2020.

\bibitem[Rueckauer et~al.(2017)Rueckauer, Lungu, Hu, Pfeiffer, and Liu]{rueckauer2017conversion}
Bodo Rueckauer, Iulia-Alexandra Lungu, Yuhuang Hu, Michael Pfeiffer, and Shih-Chii Liu.
\newblock Conversion of continuous-valued deep networks to efficient event-driven networks for image classification.
\newblock \emph{Frontiers in neuroscience}, 11:\penalty0 682, 2017.

\bibitem[Simonyan and Zisserman(2014)]{simonyan2014very}
Karen Simonyan and Andrew Zisserman.
\newblock Very deep convolutional networks for large-scale image recognition.
\newblock \emph{arXiv preprint arXiv:1409.1556}, 2014.

\bibitem[Srivastava et~al.(2014)Srivastava, Hinton, Krizhevsky, Sutskever, and Salakhutdinov]{srivastava2014dropout}
Nitish Srivastava, Geoffrey Hinton, Alex Krizhevsky, Ilya Sutskever, and Ruslan Salakhutdinov.
\newblock Dropout: a simple way to prevent neural networks from overfitting.
\newblock \emph{The journal of machine learning research}, 15\penalty0 (1):\penalty0 1929--1958, 2014.

\bibitem[Tang et~al.(2022)Tang, Lai, Xie, Yang, and Zheng]{tang2022snn2ann}
Jianxiong Tang, Jianhuang Lai, Xiaohua Xie, Lingxiao Yang, and Wei-Shi Zheng.
\newblock Snn2ann: A fast and memory-efficient training framework for spiking neural networks.
\newblock \emph{arXiv preprint arXiv:2206.09449}, 2022.

\bibitem[Touvron et~al.(2022)Touvron, Bojanowski, Caron, Cord, El-Nouby, Grave, Izacard, Joulin, Synnaeve, Verbeek, et~al.]{touvron2022resmlp}
Hugo Touvron, Piotr Bojanowski, Mathilde Caron, Matthieu Cord, Alaaeldin El-Nouby, Edouard Grave, Gautier Izacard, Armand Joulin, Gabriel Synnaeve, Jakob Verbeek, et~al.
\newblock Resmlp: Feedforward networks for image classification with data-efficient training.
\newblock \emph{IEEE Transactions on Pattern Analysis and Machine Intelligence}, 2022.

\bibitem[Van~Dyk and Meng(2001)]{van2001art}
David~A Van~Dyk and Xiao-Li Meng.
\newblock The art of data augmentation.
\newblock \emph{Journal of Computational and Graphical Statistics}, 10\penalty0 (1):\penalty0 1--50, 2001.

\bibitem[Wu et~al.(2022)Wu, Chen, and Yao]{hpcc22}
Man Wu, Zheng Chen, and Yunpeng Yao.
\newblock Learning local representation by gradient-isolated memorizing of spiking neural network.
\newblock In \emph{2022 IEEE 24th Int Conf on High Performance Computing \& Communications; (HPCC)}, pages 733--740, 2022.

\bibitem[Wu et~al.(2018)Wu, Deng, Li, Zhu, and Shi]{wu2018spatio}
Yujie Wu, Lei Deng, Guoqi Li, Jun Zhu, and Luping Shi.
\newblock Spatio-temporal backpropagation for training high-performance spiking neural networks.
\newblock \emph{Frontiers in neuroscience}, 12:\penalty0 331, 2018.

\bibitem[Wu et~al.(2019)Wu, Deng, Li, Zhu, Xie, and Shi]{wu2019direct}
Yujie Wu, Lei Deng, Guoqi Li, Jun Zhu, Yuan Xie, and Luping Shi.
\newblock Direct training for spiking neural networks: Faster, larger, better.
\newblock In \emph{Proceedings of the AAAI Conference on Artificial Intelligence}, volume~33, pages 1311--1318, 2019.

\bibitem[Wu and He(2018)]{wu2018group}
Yuxin Wu and Kaiming He.
\newblock Group normalization.
\newblock In \emph{Proceedings of the European conference on computer vision (ECCV)}, pages 3--19, 2018.

\bibitem[Yun et~al.(2019)Yun, Han, Oh, Chun, Choe, and Yoo]{yun2019cutmix}
Sangdoo Yun, Dongyoon Han, Seong~Joon Oh, Sanghyuk Chun, Junsuk Choe, and Youngjoon Yoo.
\newblock Cutmix: Regularization strategy to train strong classifiers with localizable features.
\newblock In \emph{Proceedings of the IEEE/CVF international conference on computer vision}, pages 6023--6032, 2019.

\bibitem[Zheng et~al.(2021)Zheng, Wu, Deng, Hu, and Li]{zheng2021going}
Hanle Zheng, Yujie Wu, Lei Deng, Yifan Hu, and Guoqi Li.
\newblock Going deeper with directly-trained larger spiking neural networks.
\newblock In \emph{Proceedings of the AAAI Conference on Artificial Intelligence}, volume~35, pages 11062--11070, 2021.

\end{thebibliography}
\clearpage
\title{Supplementary Material}
\maketitle
\setcounter{section}{0}

\section{Supplementary code}
The online version contains supplementary material
available at: \\
\href{https://github.com/iverss1/ECML\_SNN}{https://github.com/iverss1/ECML\_SNN}

\section{Related Work}

\subsection{Learning Methods of SNNs}

Current training methods in SNNs that achieve high performance can generally be divided into two branches: i) ANN to SNN conversion \cite{cao2015spiking} and ii) direct training with surrogate gradient. 
The conversion methods nearly maintain the accuracy of original ANNs by training the analog-spiking ANNs normally and converting them to spiking neurons by counting the fire rate \cite{rueckauer2017conversion}. 
Recent work has combined conversion and training processes to achieve near-lossless accuracy with VGG, ResNet, and their variants \cite{deng2021optimal,li2021free, han2020rmp}. 
However, the converted SNNs require longer time to rival the original ANN in precision due to pre-coding \cite{rueckauer2017conversion}, which increases the SNN's latency and restricts its practical application \cite{fang2021deep}. 
Li et al. proposed a calibration method to improve the accuracy under fewer time steps \cite{li2021free}. 
However, achieving competitive results still generally requires certain time steps (>100), which violates the hope of the low-energy costs to SNNs.
Direct training of SNNs involves surrogate gradient-descent algorithms \cite{wu2018spatio,fang2021incorporating}, as the gradient with respect to threshold-triggered firing is non-differentiable \cite{bengio2013estimating}.
To name a few, Spatio-Temporal Backpropagation \cite{wu2018spatio}, 
Explicit Iterative LIF neuron \cite{wu2019direct}, and Threshold-Dependent Batch Normalization \cite{zheng2021going}, allow gradient-based training methods to directly train SNNs using only a few time steps, such as $t=10$ \cite{rathi2020diet}, $t=6$ \cite{zheng2021going}, and unexpectedly, $t=2$ \cite{li2021differentiable} in recent research. 

\subsection{Direct Training Framework}

Recent studies have shown that incorporating advanced computational mechanisms and architectures from CNN and RNN with SNN neurons can improve performance and reduce the required time steps.
The combination of convolution kernels and spiking neurons is a main trend that enables SNNs to inherit the powerful learning ability of CNNs on local areas or points \cite{fang2021deep,li2021differentiable}.
The earliest feedforward hierarchical spiking CNN for unsupervised learning of visual features was developed by Masquelier et al. \cite{masquelier2007unsupervised}.
As SNNs have evolved, Wu et al. \cite{wu2018group} improved the leaky integrate-and-fire (LIF) model \cite{hunsberger2015spiking} to an iterative LIF model.
Li et al. \cite{li2022brain} further improved CNN-SNN by using a variant of convolution kernels, to obtain an optimal full-precision classification network.
On the other hand, RNN-SNN methods are relatively rare. 
Given the adaptability of sequence models to time series, SNNs can still handle sequence features \cite{bellec2018long}.
Recently, Rezaabad et al. \cite{lotfi2020long} developed an error backpropagation for LSTM-SNN for sequential datasets, while Datta et al. 
\cite{datta2022towards} proposed a novel activation function in the source LSTM to jointly optimize the parameters on temporal MNIST.

This paper aims to address the trade-off issue between accuracy and time step by proposing a general-purpose framework. 
The theoretical feasibility of using CNNs and RNNs in our proposal is also demonstrated.

\section{Related issues in window partition}

\subsection{Computational complexity between CNN-SNN and RNN-SNN in windows}
The computational complexity of SNN (CNN/LSTM based) within a local window on an image of $h × w$ is \ref{computational complexity of LSNN} (Regardless of the bias):
\begin{gather}
\label{computational complexity of LSNN}
\Omega( SCNN-W) =d_{x} \ *\ k_{h} \ *\ k_{w} *d_{h} *2 \notag\\
\Omega( LSNN-W) =\ d_{x} \ \ast \ d_{h} \ \ast \ 8\ +\ d_{h} \ \ast \ ( d_{h} \ \ast \ 8\ +\ 20)
\end{gather}
\begin{equation}
\label{eq4.2.2}
\frac{\Omega( SCNN-W)}{\Omega( LSNN-W)} \sim \frac{d_{x} \ *\ k_{h} \ *\ k_{w}}{\ ( d_{x} +\ d_{h}) *4}
\end{equation}
where $d_{x},d_{h}$ represent the dimension of input and output respectively,$k_{w},k_{h}$ are the size of convolution kernel.
The complexity of an algorithm is dictated by its highest order term, thus the term with complexity $O(n)$ in the $\Omega(LSNN-W)$ can be omitted.Subsequently we compared the two models and obtained \ref{eq4.2.2}.
Due to $d_{x}$ generally equals $d_{h}$ with the shortcuts in SNN to match the activations of the original input,\ref{eq4.2.2} is positive correlation according to the convolution kernel size and the kernel size usually set to $3*3$.
When we consider a LIF cell with only one layer of convolution kernel (actually more than one layer in general) and one with a layer of LSTM, it is obvious that LSNN has smaller computational complexity.

\subsection{Weighted condense algorithm in dilated window}
\renewcommand{\thealgorithm}{1} 
    \begin{algorithm}
        \caption{Recomposed Computing}  
        \begin{algorithmic}
            \Require {$Dilated\ feature\ $\{$G_1,G_2,G_3,G_4$\}, $Dilated\ window\ size \ M =\frac{3L} {4}$,\ $Window\ offset\ w_o = 2 * M - L$}\ \ \ $(L\ is\ the\ size\ of\ fearture\ map)$
            \Ensure $recomposed\ feature\ map\ X_{l}$  \\
            \textbf{Weighted-Condense:}  
            \State \ding{172}$coverG_{1} G_{2} \ =\ \frac{\left( G_{1}\left[ :,\ :,\frac{w_{o}}{2} :\frac{3w_{o}}{2} ,\ :\right] \ +\ G_{2}[ :,\ :,0:w_{o} ,\ :]\right)}{\ 2} $\ \  $\ \ \ \ \triangleright$ condense overlaps region of $G_1,G_2$
            \State \ding{173}$G_{1} G_{2} =\ cat\left(\left[ G_{1}\left[ :,\ :,\ 0:\frac{w_{o}}{2} ,\ :\right] ,coverG_{1} G_{2} ,G_{2}\left[ :,\ :,\ w_{o} :\frac{3w_{o}}{\ 2} ,\ :\right]\right]\right) $\ \  $\triangleright$ recompose $G_1,G_2$ together
            \State \textbf{Repeat:} \ding{172} $and$ \ding{173} $for\ G_{3}G_{4}$:
            \State $coverP\ =\ \ \frac{\left( G_{1} G_{2}\left[ :,\frac{w_{o}}{2} :\frac{3w_{o}}{2} ,\ :,\ :\right] \ +\ G_{3} G_{4}[ :,0:w_{o} ,:,\ :]\right)}{2}$
            \State $G\ =\ cat\left(\left[ G_{1} G_{2}\left[ :,\ 0:\frac{w_{o}}{2} ,\ :,\ :\right] ,\ coverP,G_{3} G_{4}\left[ :,\ w_{o} :\frac{3w_{o}}{2} ,\ :,\ :\right]\right]\right)$
            \State \textbf{Region Threshold:}\ $X_{l}\ =\ G.where(Ms>Th_R)\rightarrow 1$
        \end{algorithmic}
        \label{recompose}
    \end{algorithm}
Weighted condense compresses the information streams $X_l$ and $M_l$ into the original feature size with certain weights, namely $4$ and $2$.
More specifically, due to the division operation of weighted-condense, “median spikes (Ms)” (such as 0.25, 0.5, and 0.75) will be produced in overlaps region of recomposed $X_{l+1}$. 
For instance, if a spike appears in the overlapping area with weight $4$, it becomes a Ms as $0.25$.
To maintain the low power advantage of event-driven, we set a region threshold ($Th_R$) to integrate these Ms into spikes. 
This threshold is set to 0.1.

\section{Related computing process in proposed framework}

\subsection{Discrete computing process of $\partial I(t)$ and $\partial V(t)$}

\begin{equation*}
\frac{\partial I (l,t)}{\partial l} =\lim\limits _{\Delta \delta _{I}\rightarrow 1}\frac{I [\iota ,n]-I[\iota -\Delta \delta _{I} ,n]}{\Delta \delta _{I}} \ -\ \frac{\Delta \delta _{I}}{2} \cdot \frac{d^{2} I(l,t)}{dl^{2}} +O\left( \Delta \delta _{I}^{2}\right)
\end{equation*}

\begin{gather*}
\frac{\partial V(G_{m1} ,G_{m2} ,t)}{\partial G_{m1}} +\frac{\partial V(G_{m1} ,G_{m2} ,t)}{\partial G_{m2}}\\
=\lim\limits _{\Delta \delta _{m1}\rightarrow 1}\frac{V[\eta ,\upsilon ,n]-V[\eta -1,\upsilon ,n]}{\Delta \delta _{m1}} +\lim\limits _{\Delta \delta _{m2}\rightarrow 1}\frac{V[\eta ,\upsilon ,n]-V[\eta ,\upsilon -1,n]}{\Delta \delta _{m2}}\\
\ -\mathbb{H}( m_{1} ,m_{2}) +O\left( \Delta \delta _{V}^{2}\right)
\end{gather*}
Taylor Expansion of synaptic current $I(t)$ and membrane potential$V(t)$ are presented as above.
Since both functions are not changing with respect to time steps(t) and the deepening of the network is a linear change, the second derivative of $I(l,t)$ does not exist.
The accumulation of membrane potential$V(t)$ should be treated as a Taylor expansion of a multivariate composite function, and the variables change along with two directions.
Where $\mathbb{H}$ is Hessian Matrix of membrane potential.
\begin{equation*}
\mathbb{H}\left( m,m^{'}\right) =\begin{bmatrix}
\frac{\partial ^{2} f( M)}{\partial m_{1}^{2}} & \frac{\partial ^{2} f( M)}{\partial m_{1} \partial m_{2}} & \cdots  & \frac{\partial ^{2} f( M)}{\partial m_{1} \partial m_{n}}\\
\frac{\partial ^{2} f( M)}{\partial m_{2} \partial m_{1}} & \frac{\partial ^{2} f( M)}{\partial m_{2}^{2}} & \cdots  & \frac{\partial ^{2} f( M)}{\partial m_{2} \partial m_{n}}\\
\vdots  & \vdots  & \iddots  & \vdots \\
\frac{\partial ^{2} f( M)}{\partial m_{n} \partial m_{1}} & \frac{\partial ^{2} f( M)}{\partial m_{n} \partial m_{2}} & \cdots  & \frac{\partial ^{2} f( M)}{\partial m_{n}^{2}}
\end{bmatrix} \ 
\end{equation*}
As shown in the Matrix, higher order expansion composite function with $m_1,m_2$ is 0.
There are two reasons for this.
(i) Since deepening and sliding are both first order linear operation, Continuous second derivative does not exist in accumulation of membrane potential $V(t)$ with either direction.
(ii)The two variables of a multivariate function are independent.
Therefore, the joint derivative of the function is $0$.
\begin{gather}
\Omega \left(\frac{\partial V_{p} (G,t)}{\partial G}\right)\\
\Rightarrow \Omega \left(\lim\limits _{\Delta \delta _{m1}\rightarrow 1}\frac{V[\eta ,\upsilon ]-V[\eta -1,\upsilon ]}{\Delta \delta _{m1}} +\lim\limits _{\Delta \delta _{m2}\rightarrow 1}\frac{V[\eta ,\upsilon ]-V[\eta ,\upsilon -1]}{\Delta \delta _{m2}}\right)\\
\Rightarrow V[\eta ,\upsilon ]-\alpha V[\eta -1,\upsilon -1]
\label{eq5}
\end{gather}
Finally, we project the potentials in both directions onto a common space, constraining the membrane potential that ultimately affects neurons by both directional variables simultaneously as Eq. \ref{eq5}

\subsection{The results of $\nabla D$ with different $\theta$}
\begin{gather}
\mathbb{S} (\theta )=16*[sin( \frac{3\theta \pi th}{2}) *sin( \frac{\theta \pi th}{4})^2]/\left( \theta ^{2} \pi ^{2}\right) \ \ \ 
\label{eq6}
\end{gather}

To detailed analysis over-activation issue, we use the area difference $\nabla D$ ranged in $R \in (th,2*th )$ between the fusion surface ($\mathbb{S}_{Surface}=\Omega(\mathbb{X},\mathbb{Y})$) and the linear plane ($\mathbb{S}_{Plane}=\mathbb{X}+\mathbb{Y}$) as an indicator. 
This range is chosen because linear fusion only causes over-activation within this range.
The area of the fusion surface $\mathbb{S}_{Surface}$ can be simplified as Eq. \ref{eq6} which depends only on the radian factors $\theta$.

\begin{table}[]
\centering
\caption{$\nabla D$ comparison given different $\theta$.}
\label{Difference}
\newcolumntype{K}{>{\centering\arraybackslash}X}
  \begin{tabular}{*{6}{p{1.8cm}}}
    \toprule 
    \multicolumn{6}{c}{$\mathbb{S}(\theta)=16\left[\sin\left(\frac{3\theta\pi t}{2}\right) \sin\left(\frac{\theta\pi t}{4}\right)^2\right]/\left(\theta^2 \pi^2\right)$} \\
    \midrule 
    $\theta$ & 0.1 & 0.3 & 0.5 & 0.7 & 0.9 \\
    $\mathbb{S}(t=0.5)$ & 0.05 & 0.16 & 0.22 & 0.24 & 0.20 \\
    $\displaystyle \nabla D$ & $-0.31$ & $-0.21$ & $-0.14$ & $-0.13$ & $-0.17$ \\
    \bottomrule
  \end{tabular}
\end{table}

The area of the projection plane $\mathbb{S}_{Plane}$ is determined by the threshold of the LIF model, the difference $\nabla D$ depends only on the radian factors $\theta$. Theoretically, the greater the absolute difference, the greater the possibility of over-activation. 
$\mathbb{S}_{Surface}$ and $\nabla D$ are resulted in the Table. \ref{Difference} with various $\theta$, the over-activation of spiking neurons caused by membrane potential fusion can be suppressed with each $\theta$.
Different $\theta$ also lead to different levels of inhibition, when excessive activation occurs.

\section{Experimental details and additional exploration}
We modify the ResMLP and VIT architectures slightly to facilitate ANN-SNN conversion.
Patch-Merging is used for down sample, other architectures are same as \cite{touvron2022resmlp} and \cite{liu2021swin}
The architectural details are:

\textbf{ResMLP12}:
{48, F-48-shorcut , 48, PM, 96, F-96-shorcut, 96, PM, (192, F-192-shorcut)$\times$3, 384, F-384, 384, C}

\textbf{VIT12}:
{48, MLP-48 , 48, PM, 96, MLP-96, 96, PM, (192, MLP-96)$\times$3, 384, MLP-384, 384, C}
\subsection{Experiments settings}
We evaluate the performance of the proposed framework in terms of classification accuracy and inference latency on the CIFAR10 \cite{krizhevsky2009learning}, CIFAR100 \cite{krizhevsky2009learning}, and Tiny-ImageNet \cite{le2015tiny} datasets.

\subsection{Training Hyperparameters}
Standard data augmentation techniques are applied for image datasets such as padding by 4 pixels on each side, and 32 × 32 cropping by randomly sampling from the padded image or its horizontally flipped version (with 0.5 probability of flipping). 
The original 32 × 32 images are used during testing. 
Both training and testing data are normalized using channel-wise mean and standard deviation calculated from training set. 
Both SNN (CNN and RNN) are trained with cross-entropy loss with stochastic gradient descent optimization (weight decay=0.00002, momentum=0.9). 
We train the SNNs for 300 and 250 epochs for CIFAR and TinyImageNet respectively, with an initial learning rate of 0.05 and warmup learning rate is 0.001. 
The learning rate noise is limit in 0.67.
The ANNs are trained with gradient clipping rather batch-norm (BN), the Gradient clipping mode is the normal version, clip-grad is set to 20.

Additionally, dropout \cite{srivastava2014dropout} is used as the regularizer with a constant dropout mask with dropout probability=0.1 training the SNNs. 
Since mix-up \cite{yun2019cutmix} and augmentation splits \cite{van2001art} causes significant information enhancing in training, we use mixup alpha as 0.1, augmentation splits as 2-6.
During SNN training, the weights are mainly initialized using as initialization \cite{ECCV-2022-roy}. 
Upon conversion, at each training iteration with 1 time step, the SNNs are trained for 300 epochs with cross-entropy loss and adam optimizer (weight decay=0.0001). 
Initial learning rate is chosen as 0.001, which is decayed by 0.1.

\subsection{Results of other RNN-SNN model}
\begin{table}[]
    \caption{Accuracy of RNN-SNN models on CIFAR10 and Cifar100 datasets.}
    \centering
    \newcolumntype{K}{>{\centering\arraybackslash}X}
    \begin{tabular}{p{2.5cm} *{4}{p{2cm}}}
    \toprule 
    Models/Dataset & MLP & RNN & Bi-RNN & GRU \\
    \midrule 
    CIFAR10 & 84.32 & 88.32 & 89.71 & 91.66 \\
    CIFAR100 & $\times$ & 64.33 & 64.17 & 65.41 \\
    \bottomrule
   \end{tabular}
    \label{tab:my_label}
\end{table}
Table. \ref{tab:my_label} presents the results of different SNN model trained with proposed framework in RNN baseline.
Performance of LSTM-SNN are detailed analysed in main paper.
The proposed framework can still converge other RNN baseline networks, although the effect is inferior to LSTM based one.

\begin{figure}[ht]
    \centerline{\includegraphics[width=0.75\linewidth]{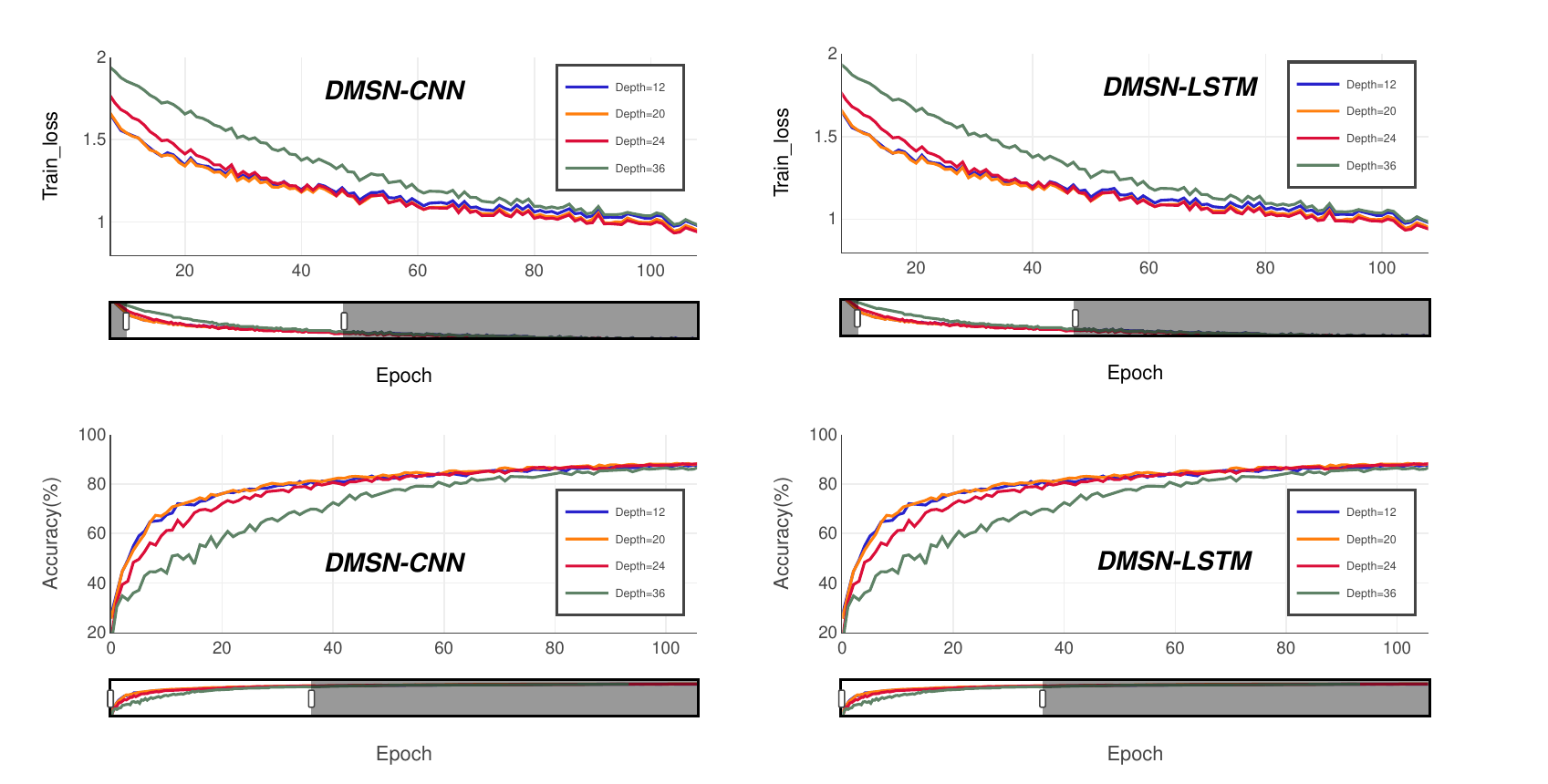}}
\caption{Deeper layers in our framework.}
\label{Deeper layers in our framework}
\end{figure}

\subsection{Training deeper network}
We also tested the convergence of our framework as the network deepened.
As shown in Fig. \ref{Deeper layers in our framework}, when the network layer number becomes 12, 20, 24, 36, the model can still be trained correctly, Whether it's based on CNN or LSTM.
And in some cases, deeper networks achieve better accuracy. However, to ensure the energy consumption of the model, we abandoned some accuracy and adopted the 12 layer network in the main paper and experiments.

\end{document}